\documentclass[twocolumn]{next++}

\usepackage[toc,page,header]{appendix}
\usepackage{minitoc}
\usepackage{amsmath,amssymb,amsfonts}
\usepackage{bm}
\usepackage{graphicx}
\usepackage{booktabs}
\usepackage{tabularx}
\usepackage{array}
\usepackage{xurl}
\usepackage{longtable}
\usepackage{multirow}
\usepackage{adjustbox}
\usepackage{wrapfig}
\usepackage{enumitem}
\usepackage{xcolor}
\usepackage{placeins}
\usepackage{flafter}
\usepackage{needspace}
\usepackage{arydshln}

\newcommand{\graycross}{\textcolor{gray}{\ensuremath{\times}}}

\title{SAVERA: Bridging Frozen Vision and Language Models via Sparse Alignment and Visual Evidence Reuse}

\author[1]{Jiajun Shen*}
\author[2]{Chuancheng Shi*}
\author[1]{Jiayi Lyu}
\author[3]{Yanpeng Sun}
\author[1]{Fei Shen$^{\dagger}$}
\author[1]{Tat-Seng Chua}

\affiliation[1]{National University of Singapore, NExT++ Research Centre}
\affiliation[2]{University of Sydney}
\affiliation[3]{Singapore Management University}

\abstract{
Independently pretrained vision and language models may learn shared structures of the same underlying world, providing a foundation for combining them to build multimodal capabilities. However, a key challenge remains in aligning cross-modal feature representations while preserving access to visual information before and after language adaptation. To this end, we propose sparse alignment with visual evidence-guided residual adaptation (SAVERA), a framework that combines frozen vision and language models through bidirectional alignment in a shared sparse space and prompt-guided residual fusion. Specifically, we first present shared-space projection, activation reconstruction and encoding (SPARE), which represents original visual, adapted visual, and textual features through shared soft-gated sparse encoding, with branch-specific reconstruction to preserve their respective information. Second, we devise bridge--text embedding and activation matching (BEAM), a bidirectional cross-modal alignment module that aligns adapted visual and textual features at both the sample and neuron levels, establishing fine-grained cross-modal correspondence. Finally, we introduce prompt-conditioned residual integration of vision and adapted features (PRIVA), a prompt-guided residual fusion module that models vision--language interactions through gating and attention and integrates complementary information from original and adapted representations through residual fusion. Extensive experiments show that SAVERA achieves competitive performance on image captioning and visual question answering while reducing object hallucination.
\vspace{-0.1cm}
}

\checkdata[Resources]{
  \href{https://github.com/Jun-Shen04-Jia/SAVERA}{\raisebox{-1ex}{\includegraphics[height=4.2ex]{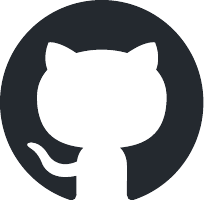}}~Code}
    \hspace{1em}\href{mailto:jiajun.shen@u.nus.edu}{\nolinkurl{jiajun.shen@u.nus.edu}}
  \hspace{1em}\href{mailto:cshi0459@uni.sydney.edu.au}{\nolinkurl{cshi0459@uni.sydney.edu.au}}
  \vspace{0.2cm}
}

\checkdata[Date]{
  November 24, 2026
}

\begin{document}
\maketitle

\begingroup
  \renewcommand{\thefootnote}{\fnsymbol{footnote}}
  \footnotetext[1]{Equal Contribution.}
  \footnotetext[2]{Corresponding Author (\href{mailto:shenfei29@nus.edu.sg}{shenfei29@nus.edu.sg}).}
\endgroup

\section{Introduction}
Independently pretrained vision and language models may learn shared structures of the same underlying world, providing a foundation for combining them to build multimodal capabilities. The Platonic Representation Hypothesis~\citep{huh2024platonic} suggests that models trained across different modalities may converge toward shared representations. Recent geometric alignment studies~\citep{schnaus2026rosetta} further show that coarse image--text correspondences can be recovered even without paired examples. However, shared representation structure does not guarantee that a language model can accurately understand and express visual semantics involving objects, actions, and relationships. Establishing fine-grained cross-modal correspondence and effectively using visual evidence to support generation therefore remain key challenges in combining independently pretrained models.

\begin{figure}[t]
  \centering
  \includegraphics[width=\linewidth]{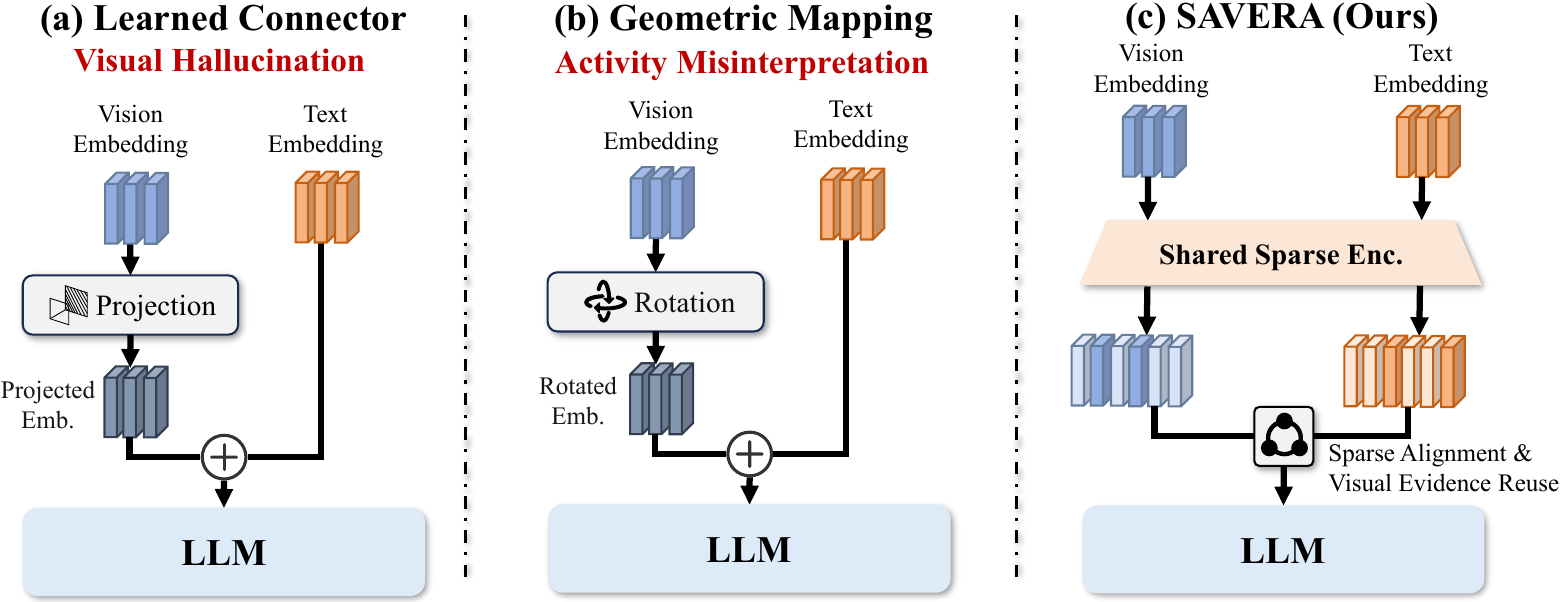}
  \vspace{-0.2cm}
  \caption{
    \textbf{Comparison of Vision--Language Interfaces.}
    (a) Learned connectors project visual features for language generation.
    (b) Geometric mappings align visual and textual representations through rotation.
    (c) SAVERA combines sparse cross-modal alignment with visual evidence reuse.
  }
  \label{fig:mapping_comparison}
  \vspace{-0.5cm}
\end{figure}

Existing methods primarily bridge vision and language through learned connectors or geometric mappings. As shown in Figure~\ref{fig:mapping_comparison}(a), learned connectors~\citep{li2023blip2,liu2024improved} project visual representations to condition a language model. Geometric mappings~\citep{schnaus2026rosetta} instead establish correspondences between visual and textual representations through transformations such as rotation, as illustrated in Figure~\ref{fig:mapping_comparison}(b). However, these methods focus on global representation matching, with limited consideration of fine-grained semantic correspondence and the effective use of information before and after adaptation. First, global correspondence does not guarantee that fine-grained information about actions and relationships is aligned at the level required for generation. Second, adaptation may change how readily object, count, and spatial cues~\citep{jiang2025finegrained,li2025lost} can be recovered from visual features, motivating the joint use of original and adapted visual information. These considerations lead to two related requirements for vision--language interfaces: establishing fine-grained cross-modal correspondence for generation and effectively using visual evidence from both original and adapted representations.

Moving from what is seen to what is said requires a clear account of the scene and evidence for the claims. Sparse representation studies~\citep{shen2025vlsae,kravets2026vista} have revealed concept-level associations between vision and language, motivating a closer examination of cross-modal correspondences involving objects, actions, and relationships. Meanwhile, adaptation reorganizes visual information for language expression, but ease of expression does not ensure that visual evidence is fully used; original representations may provide complementary cues for grounding. Multimodal generation therefore requires fine-grained semantic correspondence and the complementary use of original and adapted visual information, so that descriptions convey what is seen while remaining faithful to the visual evidence.

For that, we propose sparse alignment with visual evidence-guided residual adaptation (SAVERA). From Figure~\ref{fig:mapping_comparison}(c), SAVERA combines frozen vision and language models through bidirectional alignment in a shared sparse space and prompt-guided residual fusion. First, shared-space projection, activation reconstruction, and encoding (SPARE) uses a shared soft-gated sparse encoder to represent original visual (Vision), adapted visual (Bridge), and textual (Text) features, with branch-specific reconstruction to preserve their respective information, providing a common interface for subsequent alignment and fusion. Second, bridge--text embedding and activation matching (BEAM) bidirectionally aligns Bridge and Text at the sample and neuron levels, establishing fine-grained cross-modal correspondence. Finally, prompt-conditioned residual integration of vision and adapted features (PRIVA) models vision--language interactions through gating and attention, integrating complementary information from Vision and Bridge to produce a residual correction to the Bridge visual prefix supplied to the frozen language model. Extensive experiments show that SAVERA achieves competitive performance on image captioning and visual question answering while reducing object hallucination across the evaluated benchmark settings.
Our main contributions are as follows:

\begin{itemize}
\item We propose SAVERA, a framework that combines frozen vision and language models through a shared sparse interface while jointly addressing fine-grained cross-modal correspondence and visual evidence reuse.

\item We develop SPARE and BEAM for shared sparse encoding and bidirectional cross-modal alignment at two levels, while PRIVA reuses complementary visual evidence through prompt-guided residual fusion.

\item Extensive experiments show that SAVERA achieves competitive performance on image captioning and visual question answering while reducing object hallucination.
\end{itemize}

\section{Related Work}

\noindent\textbf{Shared Geometry and Vision--Language Interfaces.}
Research on connecting independently pretrained vision and language models~\citep{huh2024platonic,maniparambil2024vision,zhang2025alignment,schnaus2026rosetta} explores shared representational geometry and learned cross-modal interfaces. The Platonic Representation Hypothesis (PRH)~\citep{huh2024platonic} suggests that independently pretrained models may develop similar representational structures. Studies of unaligned vision and language encoders identify structural similarities~\citep{maniparambil2024vision}, while orthogonal mapping recovers coarse cross-modal correspondence without paired data~\citep{schnaus2026rosetta}. Evidence for such convergence is less stable at finer semantic granularity and under realistic many-to-many correspondences~\citep{koepke2026cave}. Unlike jointly pretrained models such as CLIP~\citep{radford2021clip}, independently pretrained vision and language models can be connected through linear mappings~\citep{merullo2023linearly} or learned interfaces, as in Flamingo~\citep{alayrac2022flamingo}, BLIP-2~\citep{li2023blip2}, InstructBLIP~\citep{dai2023instructblip}, and LLaVA~\citep{liu2023visualinstruction,liu2024improved}. Beyond connector design, aligning cross-modal feature distributions has also been studied in multimodal dataset distillation~\citep{shen2026model}. Further analyses show that adaptation can change the accessibility of object, count, and spatial information~\citep{jiang2025finegrained,li2025lost}. However, global representational compatibility alone does not establish whether fine-grained visual semantics are aligned with language and remain accessible during generation conditioned on the input image.

\noindent\textbf{Sparse Representations and Cross-Modal Concepts.}
Sparse representations~\citep{cunningham2024sparse,lim2025patchsae,shen2025vlsae,kravets2026vista} provide a feature-level perspective for analyzing concepts within and across modalities. Sparse autoencoders~\citep{cunningham2024sparse,rajamanoharan2024gated,shi2026tracerouter++,shi2026tracerouter,shi2026orthoeraser,gao2025scalingsae} decompose distributed activations into features that are easier to analyze. PatchSAE~\citep{lim2025patchsae} examines visual concepts and their changes during adaptation. For cross-modal analysis, VL-SAE~\citep{shen2025vlsae} studies concepts in a shared sparse space, while VISTA~\citep{kravets2026vista} uses textual sparse representations to interpret visual features. SAVE~\citep{park2026save} further identifies SAE-based visual understanding features to mitigate object hallucination through feature steering. Visual hallucinations have also been studied through diagnostic benchmarks~\citep{guan2024hallusionbench} and mitigation strategies targeting decoding or visual attention~\citep{huang2024opera,an2025agla,yang2025modularhallucination}, while recent work further examines whether model answers are grounded in stable latent knowledge support~\citep{shi2026correct}. Beyond sparse concept analysis, cross-modal interfaces include linear projections in MiniGPT-4~\citep{zhu2023minigpt4}, MLP connectors in LLaVA-1.5~\citep{liu2024improved}, Pixel-Unshuffle in InternVL~\citep{chen2024far}, convolutional abstractors in Honeybee~\citep{cha2024honeybee}, and learned-query resampling in BLIP-2~\citep{li2023blip2}. Orthogonal Procrustes alignment~\citep{schnaus2026rosetta} further provides a geometry-preserving alternative for aligning independently pretrained representations. However, these approaches leave open how original and language-adapted visual representations can be jointly used to support grounded generation.

\begin{figure*}[t]
    \centering
    \includegraphics[width=1\textwidth]{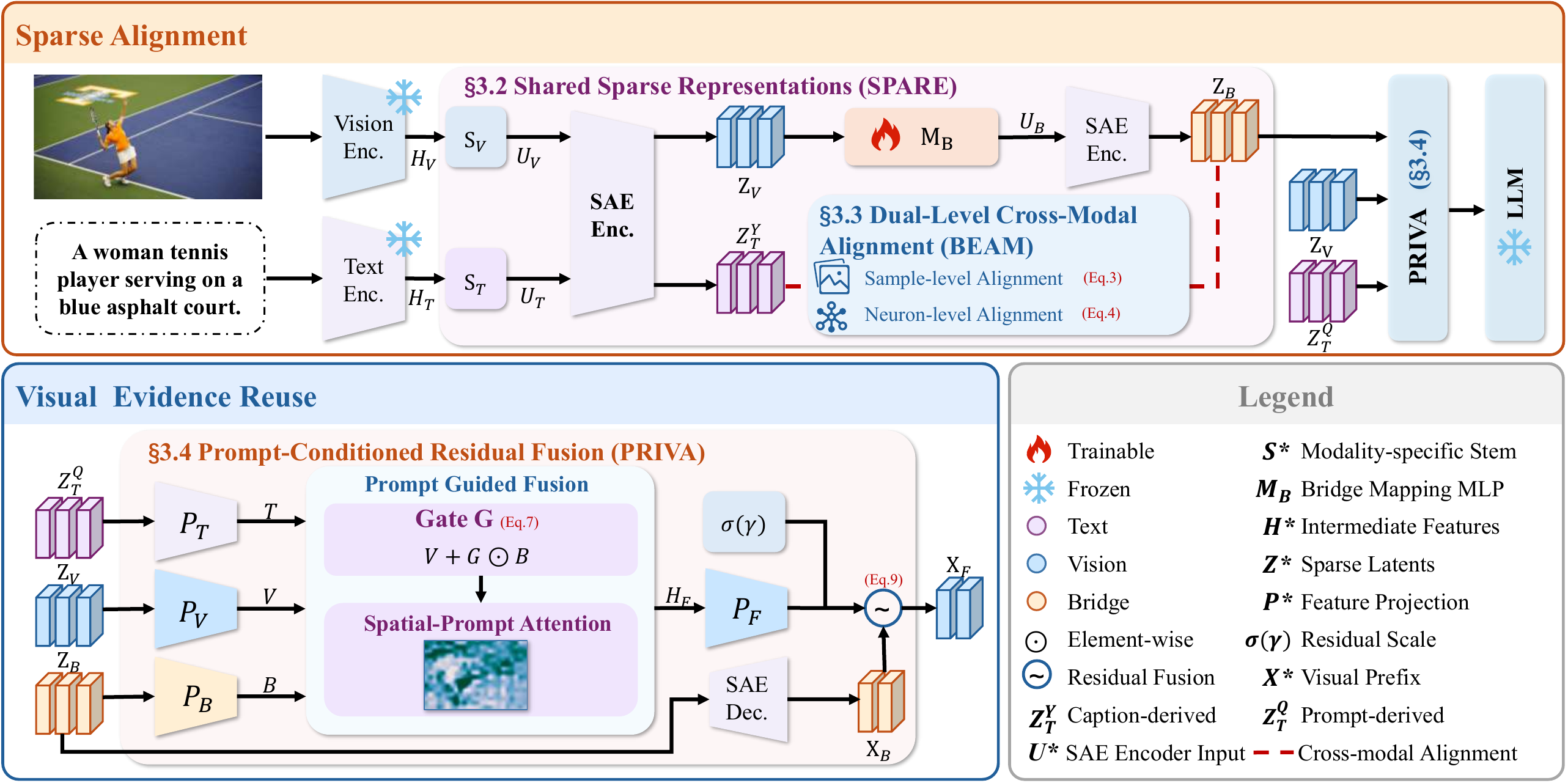}
    \vspace{-0.4cm}
    \caption{\textbf{Overview of SAVERA.} SPARE provides a shared sparse encoder for Vision, Bridge, and Text. BEAM applies the nonlinear mapping $M_B$ and aligns Bridge with caption-derived Text at the sample and neuron levels. With SPARE and BEAM frozen, PRIVA integrates Vision, Bridge, and prompt-derived Text to refine the visual prefix through residual correction.}
    \label{fig:framework}
    \vspace{-0.3cm}
\end{figure*}

\section{Method}
\label{sec:method}

\subsection{Overall Framework}
\label{sec:overview}

Given an image $I$, a prompt $Q$, and a target response $Y$, SAVERA comprises three components (Figure~\ref{fig:framework}). SPARE represents Vision, Bridge, and Text through shared sparse encoding and branch-specific reconstruction. BEAM learns a nonlinear Bridge mapping and aligns Bridge with caption-derived Text at the sample and neuron levels. PRIVA integrates original Vision, adapted Bridge, and prompt-derived Text to refine the visual prefix. The pretrained backbones remain frozen throughout. SPARE and BEAM are jointly trained and frozen before optimizing PRIVA.

\subsection{Shared Sparse Representations}
\label{sec:spare}

For an image--caption pair $(I,Y)$, frozen DINOv2 provides patch features $H_V=F_V(I)$, while frozen Vicuna provides layer-$\ell$ text states $H_T=F_T^{(\ell)}(Y)$. Modality-specific stems produce $U_V=S_V(H_V)$ and $U_T=S_T(H_T)$ in a common input width. SPARE applies a shared soft-gated sparse encoder $E$ to these inputs and the adapted visual features $U_B$ produced by BEAM (Sec.~\ref{sec:beam}):
\begin{equation}
\label{eq:encoder}
\begin{aligned}
A(U)&=\operatorname{RMSNorm}(U)W_E^\top+b_E,\\
E(U)&=\sigma\!\left(
\frac{A(U)-\boldsymbol\theta}{\tau}
\right)\odot\operatorname{softplus}(A(U)).
\end{aligned}
\end{equation}
Here $\boldsymbol\theta$ and $\tau$ control the soft gates. Activation-budget, gate-binarization, and soft-$L_0$ penalties encourage sparsity without imposing hard top-$K$ selection on the encoded features.

The same encoder represents Vision, Bridge, and Text in shared sparse coordinates, while branch-specific decoders $D_V,D_B,D_T$ reconstruct $H_V,U_B,H_T$, respectively. Reconstruction and sparsity objectives $\mathcal L_{\mathrm{rec}}$ and $\mathcal L_{\mathrm{sp}}$ regularize these representations, providing a common sparse interface while encouraging the retention of branch-specific information in the corresponding modality-specific branches.

\subsection{Dual-Level Cross-Modal Alignment}
\label{sec:beam}

Sharing sparse coordinates does not guarantee cross-modal semantic correspondence. BEAM learns a nonlinear Bridge mapping and aligns its output with caption-derived Text at two complementary levels. Given $Z_V=E(U_V)$, a learnable MLP $M_B$ produces
\begin{equation}
\label{eq:bridge_mapping}
\begin{aligned}
U_B&=M_B(Z_V),\qquad Z_B=E(U_B),\\
Z_T^Y&=E(U_T),\qquad X_B=D_B(Z_B).
\end{aligned}
\end{equation}
The Bridge representation $Z_B$ describes the same image patches as $Z_V$ after nonlinear language adaptation. The decoded output $X_B$ serves as the visual prefix to frozen Vicuna, supervised by reconstruction and autoregressive language modeling.

At the sample level, pooled $Z_B$ and $Z_T^Y$ are projected by a shared head and $\ell_2$-normalized into $b_i$ and $t_i$, respectively. Here $i$ and $j$ index paired examples and candidate negatives, while $s>0$ and $\beta$ denote the logit scale and bias. We adopt bidirectional SigLIP-style alignment~\citep{zhai2023siglip}:
\begin{equation}
\label{eq:pair_alignment}
\begin{aligned}
\ell_{B\to T}
&=\mathbb E_{i,j}
\left[\operatorname{softplus}(-y_{ij}r_{ij})\right],\\
\mathcal L_{\mathrm{pair}}
&=\frac{\lambda_{\mathrm{pair}}}{2}
\left(\ell_{B\to T}+\ell_{T\to B}\right),
\end{aligned}
\end{equation}
where $y_{ij}=+1$ for matched pairs and $-1$ otherwise. Both directions use detached cached negatives to reinforce global image--text correspondence.

At the neuron level, BEAM aligns cross-sample activation profiles of shared sparse neurons. Pooling pre-gate activations $A(U_B)$ and $A(U_T)$ over patches or valid tokens yields neuron responses across paired examples. For $q$ sampled neurons, these form matrices $R_B,R_T\in\mathbb R^{N\times q}$, whose columns capture the activation profiles of individual neurons across the same paired examples. We apply symmetric InfoNCE to match corresponding profiles and regularize their cross-modal correlation matrix, following the redundancy-reduction principle introduced by Barlow Twins~\citep{zbontar2021barlowtwins}:
\begin{equation}
\label{eq:profile_alignment}
\begin{aligned}
C&=\frac{1}{N}\widetilde R_B^\top
\widetilde R_T,\\
\mathcal L_{\mathrm{profile}}
={}&\lambda_{\mathrm{nce}}\mathcal L_{\mathrm{NCE}}
+\frac{\lambda_{\mathrm{diag}}}{q}
\sum_k(1-C_{kk})^2\\
&+\frac{\lambda_{\mathrm{off}}}{q^2}
\sum_{k\ne j}C_{kj}^2.
\end{aligned}
\end{equation}
Here $\widetilde R_B$ and $\widetilde R_T$ are standardized across examples. The contrastive term matches corresponding response profiles in both directions, while the correlation terms encourage diagonal agreement and suppress off-diagonal dependencies. Unlike sample-level matching, this objective constrains how individual sparse features respond across modalities.

SPARE and BEAM are jointly optimized through reconstruction, sparsity, language supervision, and the two complementary alignment objectives defined above:
\begin{equation}
\label{eq:adapt_objective}
\begin{aligned}
\mathcal L_{\mathrm{adapt}}={}&
\lambda_{\mathrm{LM}}\mathcal L_{\mathrm{LM}}^B
+\mathcal L_{\mathrm{rec}}
+\mathcal L_{\mathrm{sp}}\\
&+\mathcal L_{\mathrm{pair}}
+\mathcal L_{\mathrm{profile}}.
\end{aligned}
\end{equation}
Together, these objectives produce a language-compatible Bridge representation while retaining the original Vision features for subsequent fusion.

\subsection{Prompt-Conditioned Residual Fusion}
\label{sec:priva}

With SPARE and BEAM frozen, PRIVA integrates original Vision and adapted Bridge features with the available prompt. Unlike caption-derived $Z_T^Y$ used for BEAM alignment, its Text input is extracted from the prompt $Q$:
\begin{equation}
\label{eq:fusion_proj}
\begin{aligned}
Z_T^Q&=E(S_T(F_T^{(\ell)}(Q))),\\
V&=P_V(Z_V),\quad B=P_B(Z_B),\\
T&=P_T(Z_T^Q).
\end{aligned}
\end{equation}
Here $P_V$, $P_B$, and $P_T$ independently normalize and project the three sparse representations to a common fusion width. The same frozen Text branch is used during training and inference; target responses do not enter the prompt-guided fusion pathway in either of these settings.

A masked mean $\bar T=\operatorname{Pool}(T)$ conditions patch-wise feature modulation:
\begin{equation}
\label{eq:gate}
\begin{aligned}
G&=\sigma\big(g([V;B;\bar T])\big),\\
H_0&=\operatorname{LN}(V+G\odot B).
\end{aligned}
\end{equation}
The MLP $g$ predicts a gate for each patch and channel. Vision enters directly, while the gate modulates the additional Bridge contribution according to the prompt.

PRIVA then models spatial and prompt-dependent interactions among the fused visual features through self-attention, cross-attention, and feed-forward refinement:
\begin{equation}
\label{eq:attention_stack}
\begin{aligned}
H_P&=H_0+\alpha\,\mathrm{PE}_{2D},\\
H_S&=\operatorname{LN}(H_P+\operatorname{SA}(H_P)),\\
H_C&=\operatorname{LN}(H_S+\operatorname{CA}(H_S,T,T)),\\
H_F&=\operatorname{LN}(H_C+\operatorname{FFN}(H_C)).
\end{aligned}
\end{equation}
Here $\alpha$ scales the two-dimensional positional encoding. Self-attention captures interactions among image patches, while cross-attention uses visual queries and prompt-token keys and values, with text-padding masks. Thus, pooled Text conditions the initial gate, whereas individual prompt tokens guide attention.

Rather than replacing the learned Bridge prefix, PRIVA predicts a gated residual correction:
\begin{equation}
\label{eq:residual_prefix}
X_F=
\underbrace{D_B(Z_B)}_{X_B}
+\underbrace{\sigma(\gamma)P_F(H_F)}_{\Delta X}.
\end{equation}
The projection $P_F$ maps fused features to Vicuna's embedding width, and the learned scalar $\gamma$ controls the correction magnitude. Frozen Vicuna receives $[X_F;\operatorname{Embed}(Q)]$ and autoregressively predicts the target response.

Only PRIVA and its auxiliary heads are optimized:
\begin{equation}
\label{eq:priva_objective}
\begin{aligned}
\mathcal L_{\mathrm{PRIVA}}={}&
\mathcal L_{\mathrm{LM}}^F
+\lambda_{\mathrm{sem}}\mathcal L_{\mathrm{sem}}\\
&+\lambda_{\mathrm{vis}}\mathcal L_{\mathrm{vis}}
+\lambda_{\mathrm{res}}\mathcal L_{\mathrm{res}}.
\end{aligned}
\end{equation}
The answer-only language-modeling loss supervises generation. Semantic supervision aligns pooled $H_F$ with a stop-gradient, answer-derived Text teacher using cosine loss, optionally supplemented by batch contrastive alignment when multiple examples are available in the same forward pass. Visual preservation predicts frozen DINOv2 features from $H_S$, before prompt cross-attention, while residual regularization constrains $\Delta X$ relative to the frozen Bridge prefix. These objectives support residual visual reuse without replacing the language-adapted interface.

\section{Experiments}
\label{sec:experiments}

\begin{table*}[t]
    \centering
    \caption{
    \textbf{Performance on Image Captioning and Visual Question Answering Benchmarks.}
    All methods use frozen DINOv2-L/14 and Vicuna-7B-v1.5 backbones.
    Higher CIDEr, B-4, and accuracy scores are better; lower CHAIR scores are better.
    }
    \label{tab:main_results}
    \small
    \resizebox{\linewidth}{!}{
    \begin{tabular}{lcccccc}
        \toprule
        & \multicolumn{4}{c}{\textbf{Captioning}}
        & \multicolumn{2}{c}{\textbf{Visual Question Answering}} \\
        \cmidrule(lr){2-5}
        \cmidrule(l){6-7}
        Method
        & CIDEr $\uparrow$
        & B-4 $\uparrow$
        & CHAIR$_s$ $\downarrow$
        & CHAIR$_i$ $\downarrow$
        & COCO-QA Acc. $\uparrow$
        & POPE Acc. $\uparrow$ \\
        \midrule
        BLIP-2 (\textcolor{gray}{\emph{QR-64}})~\citep{li2023blip2}
        & 0.1944 & 0.0873 & 31.60 & 26.47
        & 17.16 & 61.95 \\

        Honeybee (\textcolor{gray}{\emph{C-Abstractor Lite}})~\citep{cha2024honeybee}
        & 0.2303 & 0.1033 & 32.04 & 23.40
        & 23.88 & 73.93 \\

        BLIP-2 (\textcolor{gray}{\emph{QR-32}})~\citep{li2023blip2}
        & 0.2627 & 0.1008 & 33.70 & 28.33
        & 17.74 & 59.32 \\

        MiniGPT-4 (\textcolor{gray}{\emph{Linear Projector}})~\citep{zhu2023minigpt4}
        & 0.2855 & 0.1133 & 21.64 & 16.21
        & 21.32 & 61.74 \\

        LLaVA-1.5 (\textcolor{gray}{\emph{MLP 2-layer}})~\citep{liu2024improved}
        & 0.4687 & 0.1468 & 11.26 & 8.50
        & 33.50 & 66.81 \\

        InternVL (\textcolor{gray}{\emph{Pixel-Unshuffle}})~\citep{chen2024far}
        & 0.5107 & 0.1522 & 13.84 & 10.67
        & 34.14 & 57.22 \\

        \midrule

        Orthogonal Rotation~\citep{schnaus2026rosetta}
        & 0.3546 & 0.1213 & 18.74 & 14.82
        & 25.13 & 70.29 \\

        \midrule

        \textbf{SAVERA (Ours)}
        & \textbf{0.5621} & \textbf{0.1571}
        & \textbf{9.08} & \textbf{8.15}
        & \textbf{36.07} & \textbf{82.90} \\

        \bottomrule
    \end{tabular}
    }
    \vspace{-0.2cm}
\end{table*}

\begin{figure*}[t]
    \centering
    \includegraphics[width=\linewidth]{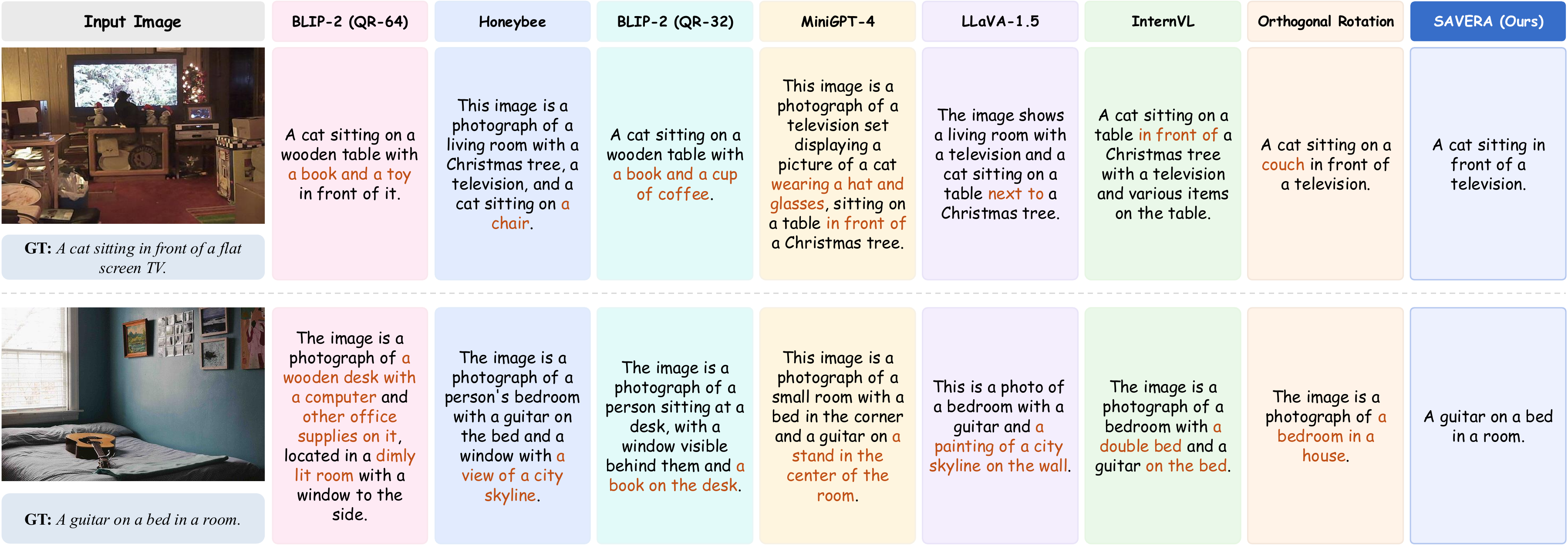}
    \caption{
        \textbf{Qualitative Comparison of Image Captioning Results on COCO.} We compare SAVERA with six representative connector baselines and orthogonal rotation on the same input images, with ground-truth (GT) captions shown below.
    }
    \label{fig:qualitative_caption}
    \vspace{-0.2cm}
\end{figure*}

\begin{figure*}[t]
    \centering
    \includegraphics[width=\linewidth]{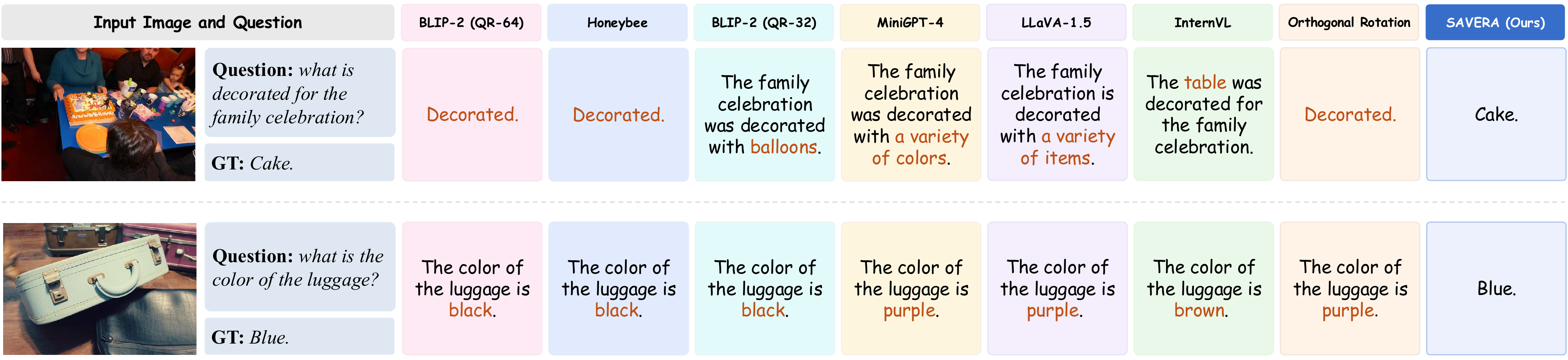}
    \caption{
        \textbf{Qualitative Comparison on COCO-QA.}
        We compare SAVERA with six connector baselines and orthogonal rotation using the same images and questions. Ground-truth (GT) answers are provided for reference.
    }
    \label{fig:qualitative_qa}
    \vspace{-0.4cm}
\end{figure*}

\subsection{Experimental Setup}
\label{sec:exp_setup}

\noindent\textbf{Models and Training.}
We use frozen DINOv2-L/14~\citep{oquab2024dinov2} and Vicuna-7B-v1.5~\citep{chiang2023vicuna}. At $336\times336$ resolution, DINOv2 produces 576 visual patch tokens. All trainable SAVERA components and connector baselines are trained exclusively on CC3M image--caption pairs~\citep{sharma2018conceptualcaptions}. SPARE and BEAM are optimized jointly, after which they are frozen for PRIVA training. PRIVA uses CC3M caption targets with answer-only language-modeling supervision and auxiliary losses. No COCO-QA or POPE annotations are used for training. We evaluate captioning on the 5,000-image MS-COCO Karpathy test split~\citep{lin2014coco,karpathy2015deepvisual}.

\noindent\textbf{Connector Baselines.}
We compare against a linear projection inspired by MiniGPT-4~\citep{zhu2023minigpt4}, a two-layer MLP following LLaVA-1.5~\citep{liu2024improved}, $2\!\times\!2$ Pixel-Unshuffle from InternVL~\citep{chen2024far}, and a lightweight convolutional abstractor inspired by Honeybee~\citep{cha2024honeybee}. We additionally evaluate BLIP-2-inspired query resamplers (QR) with 32 and 64 learned queries~\citep{li2023blip2}. We also include Orthogonal Procrustes alignment~\citep{schnaus2026rosetta}, which estimates a geometry-preserving mapping between dense visual and language representations. All baselines use the same frozen backbones and CC3M image--caption training data as SAVERA.

\noindent\textbf{Evaluation.}
We evaluate SAVERA on image captioning and visual question answering benchmarks. For captioning, we report BLEU-4 (B-4)~\citep{papineni2002bleu} and CIDEr~\citep{vedantam2015cider} to assess caption quality, alongside sentence-level $\mathrm{CHAIR}_s$ and instance-level $\mathrm{CHAIR}_i$~\citep{rohrbach2018object} to measure object hallucination. For visual question answering, we evaluate COCO-QA~\citep{ren2015imageqa} on object, number, color, and location questions, and POPE~\citep{li2023evaluating} on binary object-presence questions. COCO-QA uses case-insensitive whole-word answer matching, applied uniformly to all methods. POPE follows the official response parser, which classifies the first generated sentence as negative when an explicit negation token is present. Accuracy and CHAIR scores are reported in percent. We compare SAVERA against dense feature-based connector baselines, including orthogonal-rotation alignment. Component ablations and representation probes further examine the contributions of individual modules and visual information accessibility.

\begin{table}[t]
    \centering
    \caption{\textbf{Component Ablation on COCO Captioning.} We examine PRIVA, shared sparse encoding, and the Vision and Bridge inputs to residual fusion. Higher B-4 and CIDEr are better; lower CHAIR scores (\%) are better.}
    \vspace{-0.2cm}
    \label{tab:ablation_component}
    \resizebox{\linewidth}{!}{
    \begin{tabular}{lcccc}
        \toprule
        Method
        & CIDEr $\uparrow$
        & B-4 $\uparrow$
        & CHAIR$_s$ $\downarrow$
        & CHAIR$_i$ $\downarrow$ \\
        \midrule
        w/o PRIVA
        & 0.5009 & 0.1522 & 13.68 & 10.65 \\

        w/o Shared Sparse Encoder
        & 0.5376 & 0.1556 & 12.16 & 9.78 \\

        w/o Vision in PRIVA
        & 0.5403 & 0.1557 & 11.44 & 9.61 \\

        w/o Bridge in PRIVA
        & 0.5511 & 0.1560 & 10.27 & 9.34 \\

        \midrule

        \textbf{SAVERA (Ours)}
        & \textbf{0.5621}
        & \textbf{0.1571}
        & \textbf{9.08}
        & \textbf{8.15} \\
        \bottomrule
    \end{tabular}}
    \vspace{-0.4cm}
\end{table}

\subsection{Quantitative Comparison with SOTA Methods}
\label{sec:captioning}

\noindent\textbf{Captioning.}
To evaluate whether fine-grained cross-modal correspondence and visual evidence reuse improve image captioning while reducing object hallucination, we compare SAVERA with six representative connector baselines on the MS-COCO Karpathy test split using the same frozen vision and language backbones. As shown in Table~\ref{tab:main_results}, SAVERA achieves the highest CIDEr and B-4 scores while maintaining the lowest CHAIR$_s$ and CHAIR$_i$ among all evaluated methods. Specifically, SAVERA achieves a CIDEr of 0.5621 and a B-4 of 0.1571, outperforming the strongest CIDEr baseline, InternVL (Pixel-Unshuffle)~\citep{chen2024far}, by 0.0514 and 0.0049, respectively. Meanwhile, SAVERA reduces CHAIR$_s$ to 9.08\% and CHAIR$_i$ to 8.15\%. These results indicate that, in this setting, combining sparse alignment with prompt-guided residual fusion supports caption generation with improved semantic correspondence and fewer hallucinated object mentions.

\noindent\textbf{Visual Question Answering.}
We evaluate SAVERA on COCO-QA and POPE to examine question answering and object grounding beyond captioning. As shown in Table~\ref{tab:main_results}, SAVERA achieves the highest accuracy among the evaluated connector baselines on both benchmarks. On COCO-QA, SAVERA achieves 36.07\% accuracy, outperforming InternVL (Pixel-Unshuffle)~\citep{chen2024far} by 1.93 percentage points. On POPE, it achieves 82.90\%, exceeding Honeybee (C-Abstractor Lite)~\citep{cha2024honeybee} by 8.97 points. These results demonstrate improved question answering and object-presence verification under the evaluated settings.

\subsection{Qualitative Comparison with SOTA Methods}

\noindent\textbf{Captioning.}
To examine the differences in visual grounding across connector architectures, we compare SAVERA with representative baselines on the same COCO images. As shown in Figure~\ref{fig:qualitative_caption}, SAVERA produces concise, reference-aligned captions, while several baselines introduce additional scene details or describe different spatial relationships. Specifically, SAVERA correctly describes a cat sitting in front of a television and a guitar on a bed, whereas some baselines place the cat on a table or chair and describe the guitar as being on a stand. These examples illustrate that SAVERA can preserve salient objects and their spatial relationships while reducing object hallucination.

\noindent\textbf{Visual Question Answering.}
To examine how visual evidence reuse supports question-specific visual grounding, we compare SAVERA with representative baselines on the same COCO-QA images and questions. As shown in Figure~\ref{fig:qualitative_qa}, SAVERA provides answers consistent with the ground truth, while several baselines produce less relevant descriptions or incorrect visual attributes. Specifically, SAVERA correctly identifies the cake as the decorated object in the family celebration, whereas several baselines describe the decorations without identifying the target object. Similarly, SAVERA correctly recognizes the luggage as blue, while other methods predict black, purple, or brown. These examples illustrate that SAVERA can provide more accurate answers by leveraging visual evidence.

\begin{table}[t]
    \centering
    \caption{\textbf{Ablation of Bridge Mapping in BEAM.} We evaluate alternative Bridge mappings within BEAM by replacing the learned nonlinear mapping with linear and geometric alternatives while keeping the downstream interface fixed.}
    \vspace{-0.2cm}
    \label{tab:mapping_replace}
    \resizebox{\linewidth}{!}{
    \begin{tabular}{lccc}
        \toprule
        Mapping & CIDEr $\uparrow$ & $\mathrm{CHAIR}_s$ $\downarrow$ & Obj. recall $\uparrow$ \\
        \midrule
        Linear Replacement & 0.3220 & 23.44 & 35.08 \\
        Geometric Mapping & 0.3973 & 17.04 & 37.36 \\
        \textbf{Learned Nonlinear Mapping (Ours)} & \textbf{0.5009} & \textbf{13.68} & \textbf{38.01} \\
        \bottomrule
    \end{tabular}}
    \vspace{-0.3cm}
\end{table}

\begin{figure}[t]
    \centering
    \includegraphics[width=\linewidth]{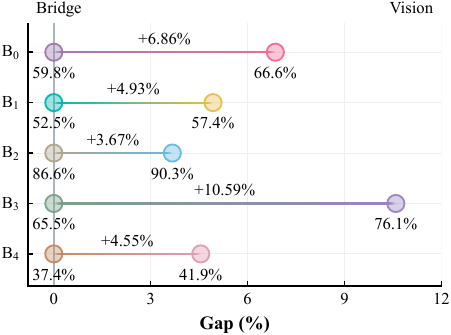}
    \vspace{-0.3cm}
    \caption{\textbf{Visual Information Accessibility Before and After Language Adaptation.} Comparison of original Vision and adapted Bridge representations across five visual diagnostics: (B0) object recognition, (B1) counting, (B2) patch localization, (B3) spatial relations, and (B4) semantic neighborhoods.}
    \label{fig:v_b}
    \vspace{-0.4cm}
\end{figure}

\begin{figure}[t]
    \centering
    \includegraphics[width=\linewidth]{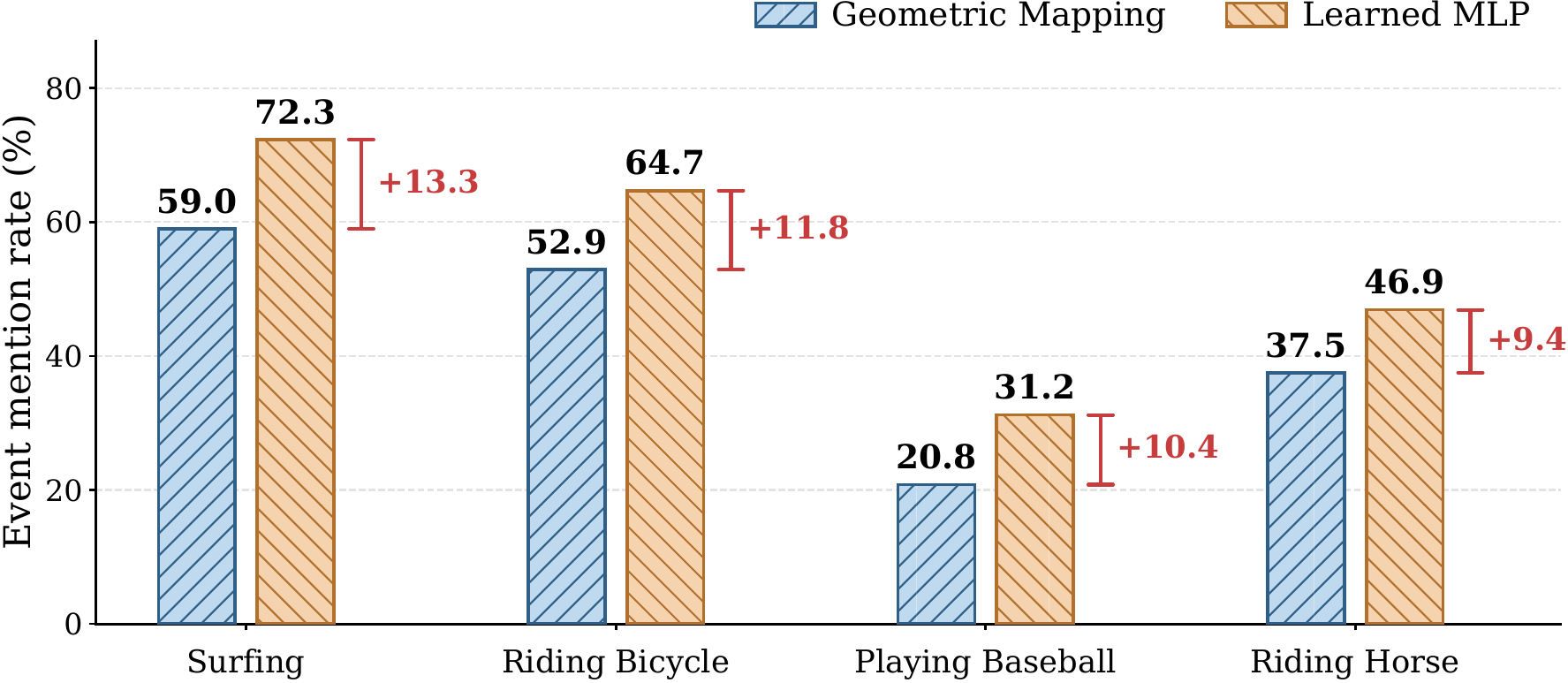}
    \caption{\textbf{Mapping-Dependent Event Expression.} Reference-event mention rates for the geometric mapping and the learned nonlinear mapping across four activities. Differences are indicated in percentage points.}
    \label{fig:mapping_event_violin}
    \vspace{-0.4cm}
\end{figure}

\subsection{Ablation Study}

\noindent\textbf{Component Ablation.}
To evaluate the contributions of sparse alignment and visual evidence reuse, we conduct an ablation study on COCO captioning by removing PRIVA, the shared sparse encoder, and Bridge features from residual fusion. As shown in Table~\ref{tab:ablation_component}, the full SAVERA achieves the highest B-4 and CIDEr scores and the lowest CHAIR$_s$ and CHAIR$_i$ across all variants. Specifically, removing PRIVA reduces CIDEr from 0.5621 to 0.5009 and increases CHAIR$_s$ from 9.08\% to 13.68\%. Removing Vision features from PRIVA reduces CIDEr to 0.5403 and increases CHAIR$_s$ to 11.44\%, while removing Bridge features yields 0.5511 CIDEr and 10.27\% CHAIR$_s$. These results indicate that prompt-conditioned residual fusion, shared sparse representations, and adapted visual features contribute to caption quality and reduce object hallucination.

\noindent\textbf{BEAM Mapping Ablation.}
To evaluate the effectiveness of nonlinear Bridge adaptation in BEAM, we conduct a mapping replacement experiment comparing the learned nonlinear mapping with linear and geometric alternatives. To ensure a fair comparison, all variants are evaluated without PRIVA, using only their adapted Bridge representations as visual prefixes for the frozen language model. As shown in Table~\ref{tab:mapping_replace}, the learned nonlinear mapping achieves the best captioning performance with the lowest object hallucination. Specifically, it achieves a CIDEr of 0.5009 and a CHAIR$_s$ of 13.68\%, compared with 0.3973 and 17.04\% for geometric mapping, and 0.3220 and 23.44\% for linear replacement. These results indicate that the nonlinear Bridge mapping in BEAM provides a more effective language-compatible visual representation.

\begin{figure}[t]
    \centering
    \includegraphics[width=1\linewidth]{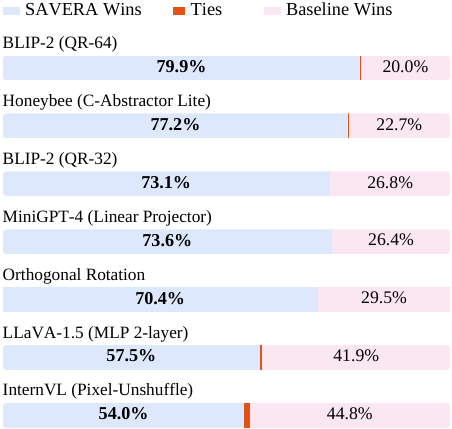}
    \vspace{-0.3cm}
    \caption{\textbf{Per-Image Captioning Comparison on the MS-COCO Karpathy Test Split.} We compare SAVERA with six connector baselines and orthogonal rotation on 5,000 images. Bars indicate the proportions of images with higher (blue), equal (red), or lower (pink) per-image CIDEr scores for SAVERA.}
    \label{fig:paired_outcomes}
    \vspace{-0.4cm}
\end{figure}

\begin{figure*}[t]
  \centering
  \includegraphics[width=1\linewidth]{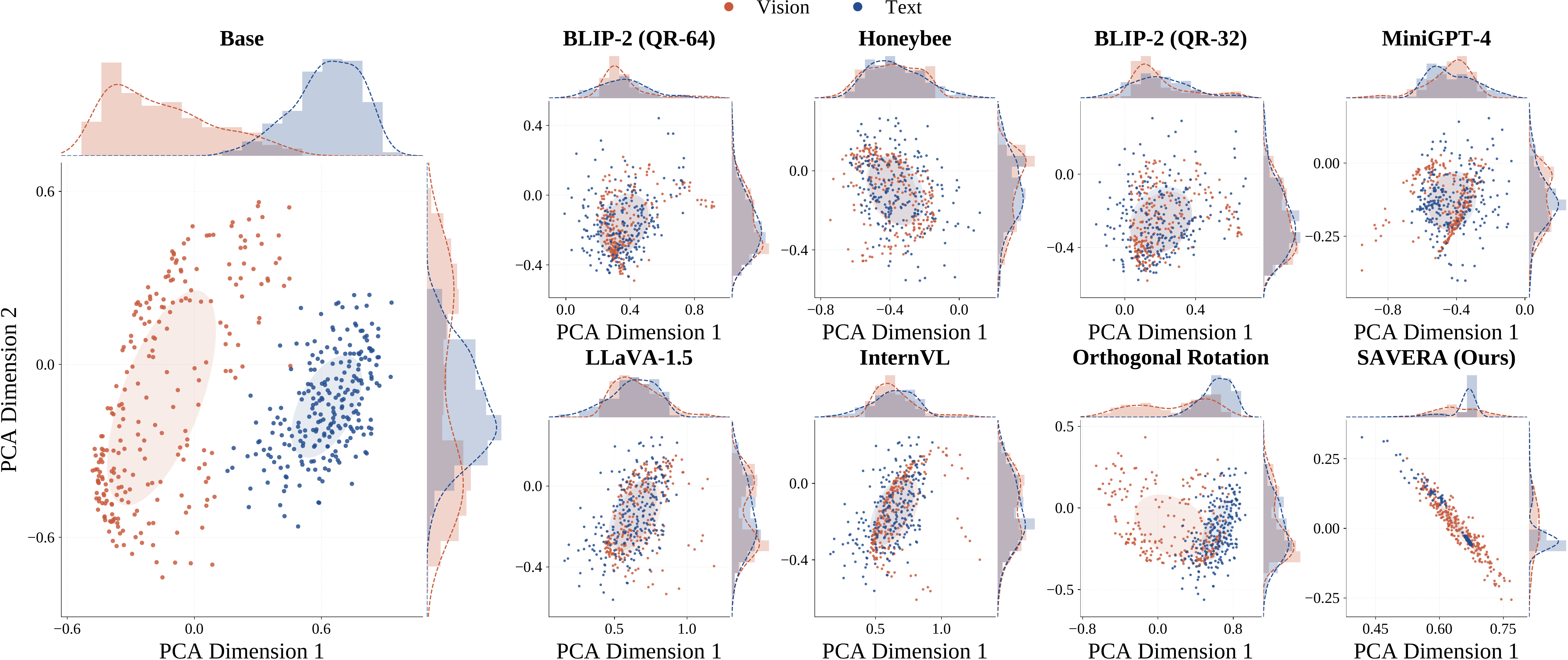}
  \caption{
    \textbf{Cross-Modal Representation Alignment.} PCA visualization of Vision (orange) and Text (blue) feature distributions for the unaligned Base, six representative connector baselines, Orthogonal Rotation, and SAVERA. Marginal density plots show the feature distributions along each PCA dimension.
  }
  \label{fig:nine}
  \vspace{-0.4cm}
\end{figure*}

\noindent\textbf{Visual Information Accessibility.}
To examine the accessibility of visual information before and after language adaptation, we compare the frozen Vision and Bridge representations across five visual diagnostics: object recognition, counting, patch localization, spatial relations, and semantic neighborhoods. As shown in Figure~\ref{fig:v_b}, Vision achieves higher scores than Bridge across all five diagnostics. Specifically, Vision achieves 76.1\% compared with 65.5\% on spatial relations and 66.6\% compared with 59.8\% on object recognition, with consistent advantages in counting, patch localization, and semantic neighborhoods. These results indicate that fine-grained visual evidence, particularly object-level and spatial information, remains more accessible in the original Vision representation, motivating its reuse alongside the language-adapted Bridge features through prompt-conditioned residual fusion.

\noindent\textbf{Event Expression Analysis.}
To examine how different alignment methods affect the expression of visual activities, we compare the reference-event mention rates of Geometric Mapping and the learned nonlinear mapping on four selected COCO activity categories. As shown in Figure~\ref{fig:mapping_event_violin}, the learned nonlinear mapping achieves higher mention rates across all four activities. Specifically, it achieves 72.3\% compared with 59.0\% for surfing, 64.7\% compared with 52.9\% for riding a bicycle, 31.2\% compared with 20.8\% for playing baseball, and 46.9\% compared with 37.5\% for riding a horse. These results suggest that learned nonlinear adaptation better supports the expression of activity-related semantic information in generated captions than the evaluated geometric mapping.

\subsection{Further Analysis}

\noindent\textbf{Per-image Captioning Comparison.}
To examine whether the improvements in caption quality extend across individual images, we conduct a per-image CIDEr comparison between SAVERA and six representative connector baselines and orthogonal rotation on the same 5,000 images from the MS-COCO Karpathy test split. As shown in Figure~\ref{fig:paired_outcomes}, SAVERA achieves higher per-image CIDEr scores than each baseline on 54.0\%--79.9\% of the evaluated images. Specifically, SAVERA outperforms InternVL (Pixel-Unshuffle) on 54.0\% of images and LLaVA-1.5 (MLP 2-layer) on 57.5\%, while achieving higher scores on 70.4\%--79.9\% of images against the remaining baselines. These results indicate that the overall CIDEr improvements are also reflected in image-level gains across different connector architectures.

\noindent\textbf{Cross-Modal Representation Alignment.}
To examine the cross-modal correspondence established by different alignment methods, we visualize the Vision and Text feature distributions using PCA. As shown in Figure~\ref{fig:nine}, the unaligned Base exhibits a clear separation between visual and textual representations, while the connector baselines show varying degrees of overlap. Specifically, SAVERA exhibits a concentrated overlap between Vision and Text features, whereas Orthogonal Rotation retains noticeable separation, and other connector baselines show more dispersed distributions. These observations suggest that shared sparse encoding and bidirectional alignment promote closer cross-modal correspondence between adapted visual and textual representations in PCA space.

\section{Conclusion}

In this paper, we proposed SAVERA, a framework that combines independently pretrained vision and language models through sparse alignment and visual evidence reuse. Specifically, SPARE establishes shared sparse representations of original visual, adapted visual, and textual features, while BEAM encourages fine-grained cross-modal correspondence through bidirectional alignment at the sample and neuron levels. PRIVA further integrates complementary information from original and adapted visual representations through prompt-guided residual fusion. Extensive experiments on image captioning and visual question answering demonstrate that SAVERA achieves competitive performance while reducing object hallucination. Further analyses show that language adaptation supports the expression of visual semantics, while original visual representations retain more accessible object-level and spatial information. These findings suggest that fine-grained cross-modal alignment and visual evidence reuse are complementary for connecting frozen vision and language models.

\bibliographystyle{plainnat}
\bibliography{main}

\clearpage

\onecolumn
\beginappendix

This appendix develops the implementation and experimental details of SAVERA beyond the main paper. We begin with a notation summary for the Vision, Bridge, and Text pathways in Section~\ref{app:notation}, followed by related studies of representation geometry, connector design, sparse features, alignment, and visual grounding in Section~\ref{app:related_work}. The complete SPARE, BEAM, and PRIVA formulations, including reconstruction, sparsity, cross-modal matching, and prompt-conditioned residual fusion, are given in Section~\ref{app:detailed_method}; model dimensions, training configurations, and evaluation protocols follow in Section~\ref{app:experimental-details}. Section~\ref{More_Experiments} then expands the benchmark comparisons with component and objective ablations, visual representation probes, controlled Vision interventions, and concept-conditioned sparse-neuron analyses. Sections~\ref{app:limitations} and~\ref{app:future_work} summarize the scope of the evaluation and directions for extension, followed by five focused questions on the design and experimental findings in Section~\ref{dis}.

\section{Notation Summary}
\label{app:notation}
Table~\ref{tab:symbols} summarizes the notation. Caption-derived $Z_T^Y$ supervises BEAM during training, whereas PRIVA uses prompt-derived $Z_T^Q$ for fusion; the target answer is not a fusion input.

\begin{table}[h]
\centering\small
\caption{Key notations in SAVERA, including the frozen DINOv2~\citep{oquab2024dinov2} and Vicuna~\citep{chiang2023vicuna} backbones.}
\label{tab:symbols}
\begin{adjustbox}{max width=\linewidth}
\begin{tabular}{@{}lp{0.77\linewidth}@{}}
\toprule
Symbol & Description \\
\midrule
$I,Q,Y$ & Input image, available instruction or question, and target response. \\
$F_V,F_T$ & Frozen visual encoder and frozen language model (text hidden-state extractor). \\
$H_V,H_T$ & DINOv2 patch features and layer-$\ell$ Vicuna text states. \\
$S_V,S_T$ & Modality-specific stems into the shared encoder input width. \\
$U_V,U_T,U_B$ & Projected vision, text, and mapped Bridge inputs to the shared sparse encoder. \\
$E,D_V,D_B,D_T$ & Shared soft-gated sparse encoder and branch-specific decoders. \\
$Z_V,Z_B,Z_T^Y,Z_T^Q$ & Sparse vision, Bridge, caption-derived text, and prompt-derived text latents. \\
$M_B$ & Trainable nonlinear Bridge mapping. \\
$A(U),\boldsymbol\theta,\tau$ & Pre-gate activations, learned neuron thresholds, and gate temperature. \\
$b_i,t_i,R_B,R_T,C$ & Sample-level normalized embeddings, neuron-profile matrices, and cross-modal correlation matrix. \\
$V,B,T,G$ & PRIVA projected features and prompt-conditioned gate. \\
$H_0,H_S,H_C,H_F$ & Gated features, self-attended features, prompt-attended features, and fused representation. \\
$X_B,\Delta X,X_F$ & Frozen Bridge prefix, learned residual correction, and final visual prefix. \\
$\sigma(\gamma)$ & Learned positive residual scale. \\
$\operatorname{sg}(\cdot)$ & Stop-gradient operation. \\
\bottomrule
\end{tabular}
\end{adjustbox}
\end{table}

\section{Additional Related Work}
\label{app:related_work}

\subsection{Representation Geometry and Independently Pretrained Models}
The relationship between independently learned visual and linguistic representations provides a motivation for connecting pretrained unimodal models. The Platonic Representation Hypothesis suggests that models trained on different modalities can exhibit related representational structure~\citep{huh2024platonic}. Analyses of unaligned vision and language encoders further investigate such similarities~\citep{maniparambil2024vision,zhang2025alignment}. Geometric approaches, including orthogonal mappings, study whether correspondences can be recovered while preserving selected aspects of representation geometry~\citep{schnaus2026rosetta}. Fine-grained correspondences between objects, activities, and attributes require resolving the multiple semantic factors and image--description relationships studied in~\citep{koepke2026cave}.

SAVERA goes beyond geometric mapping by learning a language-compatible Bridge representation and aligning it with text at the example and sparse-neuron levels. The geometric mapping comparison uses the same frozen downstream interface to examine how constrained and learned nonlinear Bridge mappings affect generation.

\subsection{Vision--Language Connectors with Frozen Backbones}
A common strategy for connecting vision and language backbones is to learn an interface between their embedding spaces. Linear projections offer a lightweight mapping~\citep{merullo2023linearly,zhu2023minigpt4}, while learned multilayer projectors are widely used in visual instruction-following systems such as LLaVA~\citep{liu2023visualinstruction,liu2024improved}. Other interfaces alter the number or organization of visual tokens. BLIP-2 uses learned queries to interact with visual features~\citep{li2023blip2}. InstructBLIP extends query-based interaction with instruction conditioning~\citep{dai2023instructblip}. Flamingo~\citep{alayrac2022flamingo} uses a learned multimodal interface; and architectures such as InternVL and Honeybee incorporate spatial rearrangement or visual abstractors~\citep{chen2024far,cha2024honeybee}. These approaches establish that different mapping and token-reduction operations can provide an effective conditioning signal for language generation.

To compare connector designs under shared frozen backbones, we implement representative projection, spatial-rearrangement, abstractor, and query-resampling operations using DINOv2 and Vicuna with CC3M supervision. Table~\ref{tab:connector_arch} specifies each operation and its visual-token budget. SAVERA combines the adapted Bridge representation with original Vision features through prompt-conditioned residual fusion.

\subsection{Sparse Autoencoders and Cross-Modal Concepts}
Sparse autoencoders (SAEs) provide a way to describe dense model activations through a larger but sparsely activated set of features~\citep{cunningham2024sparse,rajamanoharan2024gated,gao2025scalingsae}. In vision, PatchSAE analyzes fine-grained visual concepts and changes in their representations during adaptation~\citep{lim2025patchsae}. Cross-modal investigations likewise employ sparse features to examine concept associations: VL-SAE studies a unified concept representation for vision--language alignment~\citep{shen2025vlsae}, whereas VISTA examines visual semantics through text-oriented sparse representations~\citep{kravets2026vista}. Pach et al.~\citep{pach2025monosemantic} examine monosemantic features in vision--language models and show that interventions in vision-encoder sparse features can affect multimodal language-model outputs. These studies motivate a shared sparse-coordinate interface for measuring and learning cross-modal feature-response correspondence.

SPARE represents Vision, Bridge, and Text with a shared soft-gated encoder and branch-specific decoders. BEAM adds sample-level and neuron-profile alignment between Bridge and caption-derived Text. Together, the shared encoder and dual-level alignment objectives connect corresponding sparse coordinates across the adapted visual and textual representations.

\subsection{Alignment at the Example and Feature-Response Levels}
Contrastive vision--language learning, exemplified by CLIP~\citep{radford2021clip}, typically rewards agreement between embeddings of paired images and texts. SigLIP replaces the usual global softmax formulation with a pairwise sigmoid-based objective~\citep{zhai2023siglip}. Such losses establish example-level discrimination without directly constraining how individual latent dimensions respond over a collection of examples. In representation learning, correlation-based redundancy reduction, including the Barlow Twins principle~\citep{zbontar2021barlowtwins}, provides a complementary way to encourage agreement between corresponding feature responses while penalizing undesired cross-feature correlations.

BEAM combines pooled Bridge--Text matching with correspondence between pre-gate neuron-response profiles. The sample-level objective aligns paired embeddings, while the neuron-level objective aligns activation profiles across examples. Table~\ref{tab:new_beam_ablate} quantifies their individual and combined contributions.

\subsection{Visual Evidence Preservation and Hallucination Mitigation}
Language compatibility and faithful visual grounding are related but distinct requirements. Prior analyses investigate how projection or adaptation changes the accessibility of object, counting, and spatial cues in visual representations~\citep{jiang2025finegrained,li2025lost}. Studies of multimodal hallucination use dedicated evaluation procedures~\citep{guan2024hallusionbench,li2023evaluating,rohrbach2018object} or modify decoding and attention to mitigate unsupported visual claims~\citep{huang2024opera,an2025agla,yang2025modularhallucination}. Sparse-feature steering provides another direction: SAVE identifies SAE-derived visual features and intervenes on them to reduce object hallucination~\citep{park2026save}, while Hua et al.~\citep{hua2025steering} exploit SAE-derived latent directions to mitigate hallucination in existing multimodal models.

Unlike post-hoc steering, PRIVA learns prompt-conditioned fusion of original Vision, adapted Bridge, and Text features. Its residual output preserves the learned Bridge prefix. The visual probes motivate retaining both visual paths, while the controlled interventions in Tables~\ref{tab:new_global_vision} and~\ref{tab:new_local_vision} assess their contribution to generation.

\section{Additional Method Details}
\label{app:detailed_method}\label{app:theoretical-derivation}

\subsection{Overall Framework}
\label{app:full_pipeline}
Given an image $I$, an available prompt $Q$, and a target response $Y$, SAVERA connects frozen visual and language backbones using three complementary components, as described in the main paper. Shared-space Projection, Activation Reconstruction, and Encoding (SPARE) represents original visual (Vision), language-adapted visual (Bridge), and textual (Text) features in shared sparse coordinates with branch-specific reconstruction. Bridge--Text Embedding and Activation Matching (BEAM) learns the nonlinear Bridge mapping and aligns Bridge with caption-derived Text at the sample and neuron-response levels. Prompt-conditioned Residual Integration of Vision and Adapted Features (PRIVA) uses original Vision, adapted Bridge, and prompt-derived Text to refine the visual prefix through gated residual fusion. DINOv2 and Vicuna remain frozen throughout. SPARE and BEAM are jointly optimized, and their learned components are fixed when PRIVA is subsequently optimized.

For an image--caption pair $(I,Y)$, the frozen backbones yield
\begin{equation}\label{eq:app_features}
H_V=F_V(I)\in\mathbb R^{N_V\times d_V},\qquad
H_T^Y=F_T^{(\ell)}(Y)\in\mathbb R^{N_Y\times d_T},
\end{equation}
where $N_V$ is the number of image patches and $N_Y$ is the number of valid caption tokens. Modality-specific stems produce inputs to a shared sparse encoder $E:\mathbb R^{d_c}\rightarrow\mathbb R^K$. The Vision, Bridge, and Text paths are
\begin{equation}\label{eq:app_stage1_flow}
\begin{aligned}
U_V&=S_V(H_V), &Z_V&=E(U_V),\\
U_B&=M_B(Z_V), &Z_B&=E(U_B),\\
U_T^Y&=S_T(H_T^Y), &Z_T^Y&=E(U_T^Y),\\
X_B&=D_B(Z_B).&&
\end{aligned}
\end{equation}
The nonlinear mapping $M_B$ adapts Vision sparse features to a language-compatible Bridge input; $D_B$ decodes Bridge features into the visual-token prefix $X_B$ supplied to Vicuna. The number of image patches is preserved through the Vision--Bridge pathway. The shared encoder does not require identical dense representations across modalities, since $D_V$, $D_B$, and $D_T$ have branch-specific reconstruction targets.

For prompt-guided generation, the Text pathway uses the available prompt rather than the target response:
\begin{equation}\label{eq:app_prompt_text}
Z_T^Q=E\!\left(S_T(F_T^{(\ell)}(Q))\right).
\end{equation}
PRIVA combines $Z_V$, $Z_B$, and $Z_T^Q$ to predict a correction to $X_B$; the resulting prefix $X_F$ and the prompt condition the frozen language model's autoregressive response. Caption-derived $Z_T^Y$ is used to supervise BEAM during training, whereas target answers never enter PRIVA's fusion input.

The implementation uses $d_V=1024$, $d_T=d_c=4096$, sparse width $K=16384$, $N_V=576$ visual patches, and a PRIVA fusion width $d_f=512$. Accordingly, $Z_V,Z_B\in\mathbb R^{N_V\times K}$, $Z_T^Y\in\mathbb R^{N_Y\times K}$, and $Z_T^Q\in\mathbb R^{N_Q\times K}$. Image patches are pooled over spatial positions; caption and prompt states are pooled with valid-token masks. PRIVA also applies the prompt mask during text cross-attention.

\subsection{Shared Sparse Representations}
\label{app:reconstruction_sparsity}
SPARE encodes the Vision input $U_V$, the Bridge input $U_B$ constructed by BEAM, and the Text input $U_T^Y$ using the same soft-gated encoder. For any input $U$ in the common width, the encoder is
\begin{equation}\label{eq:app_shared_encoder}
\begin{aligned}
A(U)&=\operatorname{RMSNorm}(U)W_E^\top+b_E,\\
E(U)&=\sigma\!\left(\frac{A(U)-\boldsymbol\theta}{\tau}\right)\odot\operatorname{softplus}(A(U)).
\end{aligned}
\end{equation}
The threshold vector $\boldsymbol\theta$ and temperature $\tau$ control the differentiable gates; no hard top-$K$ operator is applied. Shared sparse coordinates allow the three modalities to use a common representation interface, while separate decoders preserve their respective dense feature targets.

The shared encoder $E$ has three independent linear decoders $D_V$, $D_B$, and $D_T$. Their reconstruction targets are
\begin{equation}\label{eq:app_recon_branches}
\begin{aligned}
\widehat H_V &= D_V(Z_V), & \mathcal L_{\mathrm{rec}}^V &= \mathcal D(\widehat H_V,H_V),\\
X_B &= D_B(Z_B), & \mathcal L_{\mathrm{rec}}^B &= \mathcal D(X_B,\operatorname{sg}(U_B)),\\
\widehat H_T &= D_T(Z_T^Y), & \mathcal L_{\mathrm{rec}}^T &= \mathcal D_{m_Y}(\widehat H_T,H_T).
\end{aligned}
\end{equation}
Here $m_Y$ is the valid-caption-token mask, and $\operatorname{sg}$ denotes stop-gradient. For a valid feature vector $x\in\mathbb R^d$ and target $y\in\mathbb R^d$, the reconstruction distance used for SPARE--BEAM optimization is
\begin{equation}\label{eq:app_recon_metric}
 d(x,y)=\frac{\|x-y\|_2^2}{\max(\|y\|_2^2,10^{-6})}
 +\eta_{\mathrm{cos}}\bigl(1-\cos(x,y)\bigr),\qquad \eta_{\mathrm{cos}}=0.2.
\end{equation}
The operator $\mathcal D$ averages $d$ over the applicable feature vectors; $\mathcal D_{m_Y}$ averages over valid text-token positions only. The combined objective is
\begin{equation}\label{eq:app_recon_total}
\mathcal L_{\mathrm{rec}}=\lambda_{V\mathrm{rec}}\mathcal L_{\mathrm{rec}}^V
+\lambda_{B\mathrm{rec}}\mathcal L_{\mathrm{rec}}^B
+\lambda_{T\mathrm{rec}}\mathcal L_{\mathrm{rec}}^T.
\end{equation}
The Bridge branch reconstructs its own mapped features $U_B$, not the original DINOv2 features $H_V$. Moreover, only the target argument of $\mathcal L_{\mathrm{rec}}^B$ is detached; its prediction pathway remains differentiable.

For each branch $m\in\{V,B,T\}$, let $A_m\in\mathbb R^{B\times N_m\times K}$ denote its pre-gate activations and $G_m=\sigma((A_m-\boldsymbol\theta)/\tau)$ the corresponding soft gates, where $K$ is the sparse width. We define the expected active count and the three sparsity components as
\begin{equation}\label{eq:app_sparse_terms}
\begin{aligned}
\bar k_m &= \mathbb E_{b,p}\!\left[\sum_{k=1}^K G_{m,bpk}\right],\\
\mathcal L_{\mathrm{budget}}^m &=\left(\frac{\bar k_m-K_0}{\max(K_0,1)}\right)^2,\\
\mathcal L_{\mathrm{binary}}^m &= \mathbb E_{b,p,k}[G_{m,bpk}(1-G_{m,bpk})],\\
\mathcal L_{0\text{-proxy}}^m &=\mathbb E_{b,p,k}[G_{m,bpk}].
\end{aligned}
\end{equation}
Each term is averaged across the three branches before weighting:
\begin{equation}\label{eq:app_sparse_total}
\mathcal L_{\mathrm{sp}}=\lambda_{\mathrm{budget}}\frac{1}{3}\sum_m\mathcal L_{\mathrm{budget}}^m
+\lambda_{\mathrm{binary}}\frac{1}{3}\sum_m\mathcal L_{\mathrm{binary}}^m
+\lambda_0\frac{1}{3}\sum_m\mathcal L_{0\text{-proxy}}^m.
\end{equation}
In this configuration, $K=16{,}384$, $K_0=64$, and the gate temperature follows log-linear annealing from $\tau_{\mathrm{start}}=1$ to $\tau_{\mathrm{end}}=0.1$ over the first $70\%$ of optimizer steps:
\begin{equation}\label{eq:app_temperature}
\tau(t)=\exp\!\left((1-r_t)\log\tau_{\mathrm{start}}+r_t\log\tau_{\mathrm{end}}\right),\quad
r_t=\min\!\left(\frac{t}{\max(1,\lfloor0.7T_{\mathrm{opt}}\rfloor)},1\right).
\end{equation}
The gate temperature schedule keeps the sparse encoder differentiable throughout optimization.

\subsection{Dual-Level Cross-Modal Alignment}
\label{app:alignment}
BEAM builds on the shared sparse coordinates of SPARE and couples the nonlinear Bridge mapping in Eq.~\eqref{eq:app_stage1_flow} with two complementary alignment objectives: paired-example matching of pooled Bridge/Text embeddings and cross-example matching of corresponding sparse-neuron activation profiles. The caption-derived $Z_T^Y$ is the alignment target for both objectives.

Let $W_A$ be the shared alignment projection and $m_Y$ the caption mask. For the $i$-th paired example, the normalized embeddings are
\begin{equation}\label{eq:app_pair_embeddings}
 b_i=\operatorname{norm}_2\!\left(W_A\operatorname{Pool}(Z_B^i)\right),\qquad
 t_i=\operatorname{norm}_2\!\left(W_A\operatorname{Pool}_{m_Y}(Z_T^{Y,i})\right).
\end{equation}
The current minibatch embeddings are augmented with detached historical embeddings in a first-in-first-out (FIFO) bank. For the Bridge-to-Text direction, write $\mathcal T=[t_1,\ldots,t_B,t_1^{\mathrm{bank}},\ldots,t_M^{\mathrm{bank}}]$ and define $y_{ij}=+1$ when $j=i$ is the current positive pair, and $y_{ij}=-1$ otherwise. With learnable scale $s=\min(\exp(\omega),100)$ and bias $\beta$, the implemented loss is
\begin{equation}\label{eq:app_siglip}
\begin{aligned}
\mathcal L_{B\rightarrow T}&=\frac{1}{B(B+M)}\sum_{i=1}^B\sum_{j=1}^{B+M}
\operatorname{softplus}\!\left[-y_{ij}(s\,b_i^\top \mathcal T_j+\beta)\right],\\
\mathcal L_{\mathrm{pair}}&=\tfrac{\lambda_{\mathrm{pair}}}{2}(\mathcal L_{B\rightarrow T}+\mathcal L_{T\rightarrow B}).
\end{aligned}
\end{equation}
This is the directional objective denoted by $\ell_{B\to T}$ in Eq.~\eqref{eq:pair_alignment} of the main paper, written here as $\mathcal L_{B\rightarrow T}$ to make its normalization explicit. The corresponding pairwise logit is $r_{ij}=s\,b_i^\top t_j+\beta$ for current-batch candidates and $r_{ij}=s\,b_i^\top\mathcal T_j+\beta$ when detached FIFO-bank candidates are included. The reverse term exchanges Bridge and Text; the two directions are averaged and weighted by $\lambda_{\mathrm{pair}}$. Historical embeddings serve as detached negatives, while the matching diagonal within the current minibatch defines positive pairs. The FIFO bank stores up to 512 paired embedding records and is updated after the current loss is computed.

The neuron-level objective operates on pre-gate activations rather than the sparse outputs. For paired example $i$, let
\begin{equation}\label{eq:app_pregate_pool}
 a_B^i=\operatorname{Pool}(A(U_B^i)),\qquad
 a_T^i=\operatorname{Pool}_{m_Y}(A(S_T(H_T^i))).
\end{equation}
For a sampled set of $q\le K$ coordinate indices, the columns of $R_B,R_T\in\mathbb R^{N\times q}$ contain the corresponding neuron responses across $N$ paired examples. These examples comprise the current minibatch and up to 256 detached historical pairs. The implementation selects up to half of the $q$ coordinates by current-batch activation variance and the remaining coordinates at random. The activation-variance ranking uses the current minibatch, and the subsequent profile losses incorporate up to 256 detached historical pairs as cross-example context. Let $\hat p_{B,k}$ and $\hat p_{T,k}$ denote row-wise $\ell_2$-normalized versions of the transposed matrices $R_B^\top$ and $R_T^\top$, respectively. With $S_{kj}=\hat p_{B,k}^{\top}\hat p_{T,j}/\tau_n$, symmetric profile contrastive alignment is
\begin{equation}\label{eq:app_profile_nce}
\mathcal L_{\mathrm{NCE}}=\frac12\left[
\frac1q\sum_{k=1}^q -\log\frac{e^{S_{kk}}}{\sum_j e^{S_{kj}}}
+\frac1q\sum_{k=1}^q -\log\frac{e^{S_{kk}}}{\sum_j e^{S_{jk}}}\right].
\end{equation}
For the correlation term, standardize each column of $R_B$ and $R_T$ across the $N$ examples using the population standard deviation (clamped below by $10^{-4}$). Denoting the standardized matrices by $\widetilde R_B$ and $\widetilde R_T$, form
\begin{equation}\label{eq:app_profile_corr}
\begin{aligned}
C&=\frac1N\widetilde R_B^\top\widetilde R_T,\\
\mathcal L_{\mathrm{diag}}&=\frac1q\sum_{k=1}^q(1-C_{kk})^2,\\
\mathcal L_{\mathrm{off}}&=\frac1{q^2}\sum_{k\ne j}C_{kj}^2,\\
\mathcal L_{\mathrm{profile}}&=\lambda_{\mathrm{nce}}\mathcal L_{\mathrm{NCE}}
+\lambda_{\mathrm{diag}}\mathcal L_{\mathrm{diag}}
+\lambda_{\mathrm{off}}\mathcal L_{\mathrm{off}}.
\end{aligned}
\end{equation}
If fewer than four paired examples are available across the current minibatch and the profile bank, the three profile losses are set to zero. The default profile size is $q=512$ and $\tau_n=0.1$. Historical rows serve as detached context; gradients flow through current-batch profiles.

The visual prefix $X_B=D_B(Z_B)$ is concatenated with the embeddings of the caption token sequence and passed through frozen Vicuna. Labels for all visual-prefix positions and padded text positions are ignored. Causal next-token cross-entropy on the remaining caption sequence defines $\mathcal L_{\mathrm{LM}}^B$. The training script evaluates this term every microstep by default (\texttt{lm\_every}=1); if a larger interval is used, its computed contribution is scaled by that interval. This language-modeling term contributes to the joint SPARE--BEAM objective described in the main paper.

SPARE and BEAM are optimized together with reconstruction, sparsity, image--caption alignment, neuron-profile alignment, and autoregressive caption supervision:
\begin{equation}\label{eq:app_full_stage1}
\mathcal L_{\mathrm{adapt}}=
\lambda_{\mathrm{LM}}\mathcal L_{\mathrm{LM}}^B
+\mathcal L_{\mathrm{rec}}+\mathcal L_{\mathrm{sp}}
+\mathcal L_{\mathrm{pair}}+\mathcal L_{\mathrm{profile}}.
\end{equation}
The learnable parameters belong to the modality-specific stems, shared sparse encoder, branch decoders, nonlinear Bridge mapping, and alignment heads. DINOv2 and Vicuna remain frozen, although the Bridge captioning loss propagates through Vicuna's input embeddings to the visual-prefix pathway. Cached negatives and historical activation profiles are detached; only current examples contribute gradients through these alignment losses. The cached records are updated after computing each loss.

\subsection{Prompt-Conditioned Residual Fusion}
\label{app:priva_forward}
With SPARE and BEAM frozen, PRIVA uses the original Vision representation and its adapted Bridge counterpart alongside Text features extracted from the available prompt $Q$, as in Eq.~\eqref{eq:app_prompt_text}. This is distinct from the caption-derived Text used to supervise BEAM. PRIVA projects the three streams to the same fusion width, uses pooled prompt information for patch-wise gating, and uses individual prompt tokens in cross-attention before predicting a residual correction to the learned Bridge prefix.

PRIVA uses the three frozen sparse streams but trains distinct normalization and projection heads for each one. Let $P_V$, $P_B$, and $P_T$ map from $K$ sparse dimensions to a shared fusion width $d_f$. Its projected features are
\begin{equation}\label{eq:app_priva_projection}
V=P_V(Z_V),\qquad B=P_B(Z_B),\qquad T=P_T(Z_T^Q),
\end{equation}
where each $P_*$ comprises a modality-specific layer normalization and a learned linear projection. For a prompt-token validity mask $m_Q$, PRIVA computes $\bar T=\operatorname{Pool}_{m_Q}(T)$, the masked mean of valid prompt-token features. It broadcasts $\bar T$ to the visual patches. A feature-wise gate then combines the original and adapted visual representations:
\begin{equation}\label{eq:app_priva_gate}
\begin{aligned}
G&=\sigma\!\left(g([V;B;\bar T])\right),\\
H_0&=\operatorname{LN}(V+G\odot B).
\end{aligned}
\end{equation}
Here $[\cdot;\cdot;\cdot]$ denotes concatenation along channels. The gate is a sigmoid-valued tensor in $\mathbb R^{N_V\times d_f}$, rather than a normalized probability distribution over modalities: Vision always enters the additive fusion directly, while Bridge is modulated separately at each patch and channel. The gate therefore controls the Bridge contribution independently at each visual location and channel.

A learnable scale $\alpha$ multiplies a two-dimensional sine--cosine position encoding on the $\sqrt{N_V}\times\sqrt{N_V}$ patch grid. The resulting features are refined by multi-head visual self-attention, prompt-token cross-attention, and a feed-forward layer:
\begin{equation}\label{eq:app_priva_attention}
\begin{aligned}
H_P&=H_0+\alpha\operatorname{PE}_{2D},\\
H_S&=\operatorname{LN}(H_P+\operatorname{SA}(H_P,H_P,H_P)),\\
H_C&=\operatorname{LN}(H_S+\operatorname{CA}(H_S,T,T;m_Q)),\\
H_F&=\operatorname{LN}(H_C+\operatorname{FFN}(H_C)).
\end{aligned}
\end{equation}
Text padding tokens are excluded as keys and values in the cross-attention computation. The visual-preservation auxiliary head is attached to $H_S$, \emph{before} the text cross-attention operation, so that it evaluates the spatial visual representation without using the target answer or later prompt-guided global mixing.

Finally, PRIVA projects the fused patches to the frozen language model's embedding dimension and forms
\begin{equation}\label{eq:app_priva_residual}
\Delta X=\sigma(\gamma)P_F(H_F),\qquad
X_F=X_B+\Delta X,\qquad X_B=D_B(Z_B).
\end{equation}
The learned scalar $\gamma$ controls the magnitude of the correction; $P_F$ is initialized with a small output-projection weight scale in the implementation. The Bridge prefix provides the base representation to which the learned correction is added. The frozen language model is conditioned on $[X_F;\operatorname{Embed}(Q)]$ and predicts the response autoregressively. This residual design retains the language-compatible Bridge interface while allowing the prompt-conditioned module to adjust it.

\label{app:viref_losses}
Let $Q$ and $Y$ denote the prompt and response, of token lengths $L_Q$ and $L_Y$. The frozen Vicuna receives visual-prefix embeddings $X_F$, followed by the prompt and teacher-forced answer tokens. The loss masks all visual-prefix and prompt positions and supervises only response tokens:
\begin{equation}\label{eq:app_viref_lm}
\mathcal L_{\mathrm{LM}}^F=-\frac1{L_Y}\sum_{t=1}^{L_Y}\log p_{\mathrm{Vicuna}}(y_t\mid X_F,Q,y_{<t}).
\end{equation}
The target answer never enters the PRIVA fusion input. For each CC3M image--caption pair, the captioning collator uses the fixed prompt \texttt{[Describe the image accurately and briefly]}. followed by \texttt{Answer:}, with the paired caption as the target response $Y$. All reported experiments train SPARE--BEAM and PRIVA exclusively on CC3M image--caption supervision; no additional question--answer dataset is used.

The teacher is extracted from the target answer alone (with required special tokens), using the frozen Vicuna hidden states, Text stem, shared SAE, and previously learned alignment head:
\begin{equation}\label{eq:app_semantic_teacher}
\begin{aligned}
t_Y&=\operatorname{sg}\left[\operatorname{norm}_2\!\left(
W_A\operatorname{Pool}_{m_Y}(E(S_T(F_T^{(\ell)}(Y))))\right)\right],\\
s_F&=\operatorname{norm}_2\!\left(W_{\mathrm{sem}}\operatorname{Pool}(H_F)\right).
\end{aligned}
\end{equation}
The semantic alignment loss consists of cosine distance and, when the per-forward minibatch has more than one example, symmetric batch contrastive cross-entropy:
\begin{equation}\label{eq:app_semantic_loss}
\begin{aligned}
\mathcal L_{\mathrm{cos}}&=\frac1B\sum_{i=1}^{B}(1-s_{F,i}^{\top}t_{Y,i}),\\
R_{ij}&=s_{F,i}^{\top}t_{Y,j}/0.07,\\
\mathcal L_{\mathrm{sem}}&=\begin{cases}
\mathcal L_{\mathrm{cos}},&B=1,\\
\mathcal L_{\mathrm{cos}}+0.125\,[\operatorname{CE}(R,\boldsymbol\iota)+\operatorname{CE}(R^\top,\boldsymbol\iota)],&B>1,
\end{cases}
\end{aligned}
\end{equation}
Here $\boldsymbol\iota=(1,\ldots,B)$ denotes the matching diagonal indices, $\operatorname{CE}(R,\boldsymbol\iota)$ is row-wise cross-entropy, and $0.125=0.25/2$. The teacher is used only for training. In the reported two-example microbatch configuration ($B=2$), the cosine term and symmetric batch contrastive cross-entropy are both evaluated. The contrastive term is formed within each forward minibatch, independently of gradient accumulation.

A linear auxiliary head predicts the frozen DINOv2 patch features from $H_S$, the spatial-self-attention output \emph{before} prompt cross-attention:
\begin{equation}\label{eq:app_visual_loss}
\begin{aligned}
\widehat H_V&=P_{\mathrm{vis}}(H_S),\\
\mathcal L_{\mathrm{vis}}&=
\frac{\mathbb E[(\widehat H_V-H_V)^2]}{\max(\mathbb E[H_V^2],10^{-6})}
+0.2\,\mathbb E_p[1-\cos(\widehat H_{V,p},H_{V,p})].
\end{aligned}
\end{equation}
The expectations in the first ratio are taken over all batch, patch, and channel elements. Unlike the SPARE reconstruction loss in Eq.~\eqref{eq:app_recon_metric}, this objective normalizes the \emph{global mean squared error} by the global mean squared target magnitude.

For $\Delta X=\sigma(\gamma)P_F(H_F)$ and the frozen Bridge prefix $X_B=D_B(Z_B)$, the implemented penalty is
\begin{equation}\label{eq:app_delta_loss}
\mathcal L_{\mathrm{res}}=\frac{\mathbb E[(\Delta X)^2]}
{\max(\mathbb E[(\operatorname{sg}(X_B))^2],10^{-6})}.
\end{equation}
It penalizes the \emph{scaled} residual relative to the frozen prefix. Together with Eq.~\eqref{eq:app_viref_lm}--\eqref{eq:app_delta_loss}, these terms form the PRIVA objective described in the main paper. Neither the semantic teacher nor the visual prediction head is required at inference.

The complete objective combines answer-only language-modeling supervision with semantic, visual-preservation, and residual-magnitude terms:
\begin{equation}\label{eq:app_full_stage2}
\mathcal L_{\mathrm{PRIVA}}=
\mathcal L_{\mathrm{LM}}^F
+\lambda_{\mathrm{sem}}\mathcal L_{\mathrm{sem}}
+\lambda_{\mathrm{vis}}\mathcal L_{\mathrm{vis}}
+\lambda_{\mathrm{res}}\mathcal L_{\mathrm{res}}.
\end{equation}
The three sparse-feature projection heads, gate, visual self-attention, text cross-attention, feed-forward block, output projection, residual scale, and auxiliary heads are trainable; SPARE, BEAM, and both pretrained backbones remain fixed. The target answer supplies teacher-forced response labels and a separate stop-gradient semantic teacher, but is never used to construct $Z_T^Q$ or the PRIVA fusion input. At the reported microbatch size $B=2$, semantic supervision uses both cosine alignment and symmetric batch contrastive cross-entropy.

At inference, SAVERA extracts $Z_V$, $Z_B$, $X_B$, and prompt-derived $Z_T^Q$, evaluates the PRIVA fusion and residual correction, and generates responses from the frozen Vicuna. No ground-truth caption, target-answer teacher, reconstruction loss, or auxiliary prediction head is needed. The optimization and evaluation protocols described here correspond to the main SAVERA implementation. The frozen mapping-replacement comparisons use their own controlled evaluation protocol.

\section{Detailed Experimental Setup}
\label{app:experimental-details}

\begin{table}[t]
\centering\small
\caption{Optimization settings for SPARE--BEAM and PRIVA, trained on CC3M~\citep{sharma2018conceptual} using AdamW~\citep{loshchilov2019decoupled}.}\label{tab:app_hyperparams}
\begin{adjustbox}{max width=\linewidth}
\begin{tabular}{@{}lcc@{}}\toprule
Setting & SPARE + BEAM & PRIVA\\\midrule
Optimizer & AdamW & AdamW\\
Learning Rate & $2\times10^{-4}$ & $1\times10^{-4}$\\
Weight Decay & $0.01$ & $0.01$\\
Microbatch Size / Accumulation & $2$ / $4$ & $2$ / $4$\\
Sparse Temperature & $1.0\rightarrow0.1$ & $0.1$ (frozen SAE)\\
$\lambda_{V\mathrm{rec}},\lambda_{B\mathrm{rec}},\lambda_{T\mathrm{rec}}$ & $1.0,0.25,1.0$ & --\\
$\lambda_{\mathrm{budget}},\lambda_{\mathrm{binary}},\lambda_0$ & $1.0,0.05,0.01$ & --\\
$\lambda_{\mathrm{pair}}$ & $0.20$ & --\\
$\lambda_{\mathrm{nce}},\lambda_{\mathrm{diag}},\lambda_{\mathrm{off}}$ & $0.10,0.05,0.01$ & --\\
$\lambda_{\mathrm{LM}}$ & $1.0$ & $1.0$ (implicit)\\
$\lambda_{\mathrm{sem}},\lambda_{\mathrm{vis}},\lambda_{\mathrm{res}}$ & -- & $0.05,0.05,0.02$\\
\bottomrule
\end{tabular}
\end{adjustbox}
\end{table}

\subsection{Backbones, Network Architecture, and Training Configuration}
\label{app:architecture_training}
We use the pretrained DINOv2-L/14 visual encoder~\citep{oquab2024dinov2} and Vicuna-7B-v1.5 language model~\citep{chiang2023vicuna}, keeping both backbones frozen throughout training. The following specifications describe the extracted representations and the trainable SPARE, BEAM, and PRIVA modules.

DINOv2-L/14~\citep{oquab2024dinov2} provides patch features at $336\times336$ resolution, yielding $24\times24=576$ patch tokens of width 1024 after excluding the class token. Vicuna-7B-v1.5~\citep{chiang2023vicuna} supplies 4096-dimensional text states from hidden-state index $\ell=16$; the DINOv2 features are read from index $-2$. The shared SAE has canonical input width 4096 and sparse width 16,384. The Vision stem is RMSNorm followed by a $1024\rightarrow4096$ linear layer. The Text stem is RMSNorm plus a learnably scaled $4096\rightarrow512\rightarrow4096$ residual adapter. The Bridge mapping is RMSNorm followed by $16384\rightarrow2048\rightarrow4096$ linear layers with SiLU between them. Each reconstruction decoder is an independent linear layer, and the shared Bridge--Text alignment head maps $16384\rightarrow1024$.

The PRIVA projections independently normalize and map each sparse representation from $16384$ to a fusion width of 512. The prompt-conditioned gate uses a $1536\rightarrow512\rightarrow512$ MLP with SiLU, producing a separate sigmoid gate for every image patch and feature channel. PRIVA uses an 8-head spatial self-attention layer followed by an 8-head prompt cross-attention layer; thus each head has width 64. A $512\rightarrow2048\rightarrow512$ GELU feed-forward network precedes the output projection $P_F:512\rightarrow4096$. The visual prediction and semantic heads map $512\rightarrow1024$. The initial residual logit is $\gamma=-3$ and the initial position-encoding scale is $\alpha=0.1$. All SPARE and BEAM components and both pretrained backbones are frozen during this stage.

To document the two-stage optimization, Table~\ref{tab:app_hyperparams} reports the AdamW~\citep{loshchilov2019decoupled} settings, microbatch and accumulation configurations, sparse-gate schedule, and loss weights used for SPARE--BEAM and PRIVA.

Both stages use a microbatch size of $2$ and gradient accumulation over $4$ microsteps, so that each optimizer update aggregates four microbatches. SPARE--BEAM optimization uses a 3\% warmup followed by cosine learning-rate decay and default BF16 mixed precision; the DINOv2 backbone is evaluated in FP32. PRIVA training is configured with cosine learning-rate decay, gradient clipping at norm 1, a maximum of 2,500 optimizer steps, and a 3\% validation split. Both SPARE--BEAM and PRIVA are trained exclusively on CC3M image--caption pairs~\citep{sharma2018conceptual}, without additional question--answer training data. Image preprocessing pads each image to a square before resizing it to $336\times336$, using the image processor's channel-mean color for padding.

\begin{table}[t]
\centering
\small
\setlength{\tabcolsep}{5pt}
\caption{Connector architectures and visual token budgets. The connector names follow Table~\ref{tab:main_results} of the main paper; references identify the source architectures.}
\label{tab:connector_arch}
\begin{adjustbox}{max width=\linewidth}
\begin{tabular}{@{}llc@{}}
\toprule
Connector & Main Operation & Visual Tokens \\
\midrule
MiniGPT-4 (\textcolor{gray}{Linear Projector})~\citep{zhu2023minigpt4} & Single Linear Projection & 576 \\
LLaVA-1.5 (\textcolor{gray}{MLP 2-layer})~\citep{liu2024improved} & Two-Layer MLP with GELU & 576 \\
InternVL (\textcolor{gray}{Pixel-Unshuffle})~\citep{chen2024far} & $2\times2$ Spatial Rearrangement Followed by MLP & 144 \\
Honeybee (\textcolor{gray}{C-Abstractor Lite})~\citep{cha2024honeybee} & Convolutional Processing and Spatial Pooling & 144 \\
BLIP-2 (\textcolor{gray}{QR-32})~\citep{li2023blip2} & 32 Learned Queries with Attention & 32 \\
BLIP-2 (\textcolor{gray}{QR-64})~\citep{li2023blip2} & 64 Learned Queries with Attention & 64 \\
\textbf{SAVERA (Ours)} & Prompt-Conditioned Fusion and Residual Decoding & 576 \\
\bottomrule
\end{tabular}
\end{adjustbox}
\end{table}

\Needspace{10\baselineskip}
\subsection{Connector Baselines and Token Budgets}
\label{app:baselines}

To compare connector architectures, we implement linear projection~\citep{zhu2023minigpt4}, two-layer MLP projection~\citep{liu2024improved}, pixel-unshuffle projection~\citep{chen2024far}, convolutional abstraction~\citep{cha2024honeybee}, and learned-query resampling~\citep{li2023blip2} with the same frozen DINOv2 and Vicuna backbones. Each design is trained with the same CC3M image--caption supervision.

The token budgets in Table~\ref{tab:connector_arch} reflect the operations of each connector: MiniGPT-4 (Linear Projector) and LLaVA-1.5 (MLP 2-layer) retain all 576 visual patches, InternVL and Honeybee use 144 visual tokens, and BLIP-2 query resampling uses 32 or 64 tokens. The resulting benchmark therefore compares these architectures at their respective visual-token budgets.

\subsection{Datasets and Evaluation Protocols}
\label{app:datasets}

We evaluate image captioning on the 5,000-image MS-COCO~\citep{lin2014coco} Karpathy test split~\citep{karpathy2015deepvisual} using BLEU-1--4, ROUGE-L, and CIDEr. We also report annotation-based CHAIR$_s$ and CHAIR$_i$~\citep{rohrbach2018object}, measuring hallucinated object mentions at the caption and object-mention levels.

The representation probes use 12,000 COCO train2014 images for training, 2,000 disjoint train2014 images for validation, and 5,000 val2014 images for testing. Their task-specific metrics are described in Appendix~\ref{app:accessibility}.

The BEAM mapping ablation replaces the Bridge mapping while keeping PRIVA disabled and the downstream interface fixed. Lower CHAIR scores indicate better hallucination performance; other reported scores are higher-is-better unless stated otherwise.

All trainable connectors use frozen DINOv2-L/14 and Vicuna-7B-v1.5 with CC3M image--caption supervision~\citep{sharma2018conceptual}. COCO-QA~\citep{ren2015imageqa} is evaluated using case-insensitive whole-word answer matching; POPE~\citep{li2023evaluating} uses the same response parser across methods. Neither benchmark supplies training annotations. Aggregate results and question-type scores are reported in Table~\ref{tab:main_qa}.

\section{More Experiments}
\label{More_Experiments}

\begin{table}[t]
\centering
\small
\setlength{\tabcolsep}{4pt}
\caption{Complete captioning results on the 5,000-image MS-COCO~\citep{lin2014coco} Karpathy test split~\citep{karpathy2015deepvisual}. Higher BLEU, ROUGE-L, and CIDEr scores are better; CHAIR$_s$ and CHAIR$_i$~\citep{rohrbach2018object} are percentages (lower is better), consistent with the main paper.}
\label{tab:complete_captioning}
\begin{adjustbox}{max width=\linewidth}
\begin{tabular}{@{}lcccccccc@{}}
\toprule
Method & B-1 & B-2 & B-3 & B-4 & R-L & CIDEr & CHAIR$_s$ & CHAIR$_i$ \\
\midrule
MiniGPT-4 (\textcolor{gray}{Linear Projector})~\citep{zhu2023minigpt4} & 0.4407 & 0.2907 & 0.1820 & 0.1133 & 0.3915 & 0.2855 & 21.64 & 16.21 \\
LLaVA-1.5 (\textcolor{gray}{MLP 2-layer})~\citep{liu2024improved} & 0.5180 & 0.3541 & 0.2295 & 0.1468 & 0.4311 & 0.4687 & 11.26 & 8.50 \\
InternVL (\textcolor{gray}{Pixel-Unshuffle})~\citep{chen2024far} & 0.5410 & 0.3695 & 0.2383 & 0.1522 & 0.4392 & 0.5107 & 13.84 & 10.67 \\
Honeybee (\textcolor{gray}{C-Abstractor Lite})~\citep{cha2024honeybee} & 0.4188 & 0.2716 & 0.1669 & 0.1033 & 0.3742 & 0.2303 & 32.04 & 23.40 \\
BLIP-2 (\textcolor{gray}{QR-32})~\citep{li2023blip2} & 0.4496 & 0.2813 & 0.1678 & 0.1008 & 0.3788 & 0.2627 & 33.70 & 28.33 \\
BLIP-2 (\textcolor{gray}{QR-64})~\citep{li2023blip2} & 0.4008 & 0.2477 & 0.1465 & 0.0873 & 0.3560 & 0.1944 & 31.60 & 26.47 \\
Orthogonal Rotation~\citep{schnaus2026rosetta} & 0.4816 & 0.3024 & 0.1936 & 0.1213 & 0.3814 & 0.3546 & 18.74 & 14.82 \\
\textbf{SAVERA (Ours)} & \textbf{0.5745} & \textbf{0.3859} & \textbf{0.2476} & \textbf{0.1571} & \textbf{0.4430} & \textbf{0.5621} & \textbf{9.08} & \textbf{8.15} \\
\bottomrule
\end{tabular}
\end{adjustbox}
\end{table}

The additional experiments proceed from end-task performance to the internal mechanisms of SAVERA. Complete captioning metrics, COCO-QA categories, and POPE sampling comparisons appear in Tables~\ref{tab:complete_captioning}--\ref{tab:new_qa_breakdown}. Tables~\ref{tab:component_appendix}--\ref{tab:new_fusion_full} then examine the contributions of shared sparse encoding, BEAM alignment, and PRIVA through controlled replacements and ablations. Finally, the visual probes, global and localized interventions, and concept-conditioned neuron analyses in Tables~\ref{tab:five_task}--\ref{tab:new_neuron_gen} characterize how original visual evidence and sparse cross-modal associations affect representations and generation.

\subsection{Complete Image Captioning Metrics}
\label{app:complete_captioning}

The complete captioning evaluation extends the main-paper comparison to BLEU-1--3 and ROUGE-L on the 5,000-image MS-COCO Karpathy test split, alongside CIDEr and CHAIR. Across the connector designs and Orthogonal Rotation in Table~\ref{tab:complete_captioning}, SAVERA achieves the highest BLEU scores at all four orders, with CIDEr 0.5621, B-4 0.1571, and ROUGE-L 0.4430; the strongest CIDEr baseline, InternVL (Pixel-Unshuffle), reaches 0.5107, 0.1522, and 0.4392, respectively. SAVERA's CHAIR$_s$ and CHAIR$_i$ are 9.08\% and 8.15\%. Relative to the Bridge-only SPARE--BEAM model in Table~\ref{tab:new_beam_ablate}, adding PRIVA improves CIDEr by 0.0612 while reducing CHAIR$_s$ and CHAIR$_i$ by 4.60 and 2.50 percentage points. The joint gains in caption overlap and hallucination metrics demonstrate the contribution of prompt-guided visual evidence reuse to grounded generation.

\begin{table}[t]
\centering\small
\caption{COCO-QA~\citep{ren2015imageqa} category and overall accuracy, and POPE~\citep{li2023evaluating} accuracy (\%). Connector names denote implementations inspired by the cited architectures under shared frozen backbones.}
\label{tab:main_qa}
\begin{adjustbox}{max width=\linewidth}
\begin{tabular}{@{}lcccccc@{}}\toprule
Method & Object & Number & Color & Location & COCO-QA & POPE \\ \midrule
BLIP-2 (\textcolor{gray}{QR-64})~\citep{li2023blip2} & 15.72 & 7.48 & 26.62 & 18.85 & 17.16 & 61.95 \\
Honeybee (\textcolor{gray}{C-Abstractor Lite})~\citep{cha2024honeybee} & 24.74 & 5.01 & 28.05 & 24.50 & 23.88 & 73.93 \\
BLIP-2 (\textcolor{gray}{QR-32})~\citep{li2023blip2} & 16.75 & 8.75 & 25.49 & 18.20 & 17.74 & 59.32 \\
MiniGPT-4 (\textcolor{gray}{Linear Projector})~\citep{zhu2023minigpt4} & 21.30 & 7.04 & 26.23 & 24.62 & 21.32 & 61.74 \\
LLaVA-1.5 (\textcolor{gray}{MLP 2-layer})~\citep{liu2024improved} & 34.95 & 8.98 & 38.19 & 32.50 & 33.50 & 66.81 \\
InternVL (\textcolor{gray}{Pixel-Unshuffle})~\citep{chen2024far} & 37.32 & 10.49 & 31.05 & 33.74 & 34.14 & 57.22 \\
Orthogonal Rotation~\citep{schnaus2026rosetta} & 26.03 & 5.27 & 29.52 & 25.78 & 25.13 & 70.29 \\
\textbf{SAVERA (Ours)} & 38.15 & 11.03 & 38.62 & 34.41 & 36.07 & 82.90 \\
\bottomrule
\end{tabular}
\end{adjustbox}
\end{table}

\begin{table}[t]
\centering\small
\caption{POPE~\citep{li2023evaluating} object-presence accuracy under random, popular, and adversarial negative sampling (\%).}
\label{tab:new_qa_breakdown}
\begin{adjustbox}{max width=\linewidth}
\begin{tabular}{@{}lcccc@{}}\toprule
Method & Random & Popular & Adversarial & Average \\ \midrule
InternVL (\textcolor{gray}{Pixel-Unshuffle})~\citep{chen2024far} & 61.00 & 58.10 & 52.57 & 57.22 \\
w/o PRIVA & 77.50 & 75.10 & 72.30 & 74.97 \\
\textbf{SAVERA (Ours)} & \textbf{84.27} & \textbf{82.83} & \textbf{81.60} & \textbf{82.90} \\
\bottomrule
\end{tabular}
\end{adjustbox}
\end{table}

\subsection{COCO-QA and POPE Evaluation}
\label{app:qa_primary}

We further examine whether the improvements extend beyond caption generation by evaluating COCO-QA and POPE using the same connector configurations. The category-level breakdown in Table~\ref{tab:main_qa} shows that SAVERA reaches 38.15\% Object, 11.03\% Number, 38.62\% Color, and 34.41\% Location accuracy, giving an overall COCO-QA score of 36.07\%. This exceeds InternVL (Pixel-Unshuffle), the strongest baseline on overall COCO-QA (34.14\%), by 1.93 percentage points. On POPE, SAVERA obtains 82.90\% accuracy, surpassing Honeybee (C-Abstractor Lite), the strongest comparator at 73.93\%, by 8.97 points. Improvements across question types and object-presence judgments indicate that the adapted Bridge and reused visual evidence together support a broader range of visual grounding tasks.

\subsection{POPE Evaluation by Negative-Sampling Strategy}
\label{app:qa_breakdown}

Object-presence judgments are also examined under the random, popular, and adversarial negative-sampling strategies of POPE, using a common decoding and response parser. SAVERA achieves 84.27\%, 82.83\%, and 81.60\% accuracy across the three conditions in Table~\ref{tab:new_qa_breakdown}, compared with 77.50\%, 75.10\%, and 72.30\% for SPARE--BEAM (w/o PRIVA). InternVL (Pixel-Unshuffle) scores 61.00\%, 58.10\%, and 52.57\% under the same conditions. SAVERA's three-way average is 82.90\%, consistent with the aggregate POPE score in Table~\ref{tab:main_qa}; its advantage is maintained across all three negative-sampling strategies.

\begin{table}[t]
\centering\small
\caption{Component ablation on COCO~\citep{lin2014coco} captioning, evaluated with CHAIR~\citep{rohrbach2018object}. Higher CIDEr and B-4 are better, whereas lower CHAIR scores are better; CHAIR entries are percentages. The ``w/o Self + Cross Attention'' configuration removes both visual self-attention and prompt cross-attention, compared with single-block ablations in Table~\ref{tab:new_priva_arch}.}
\label{tab:component_appendix}
\begin{adjustbox}{max width=\linewidth}
\begin{tabular}{@{}lcccc@{}}
\toprule
Variant & CIDEr $\uparrow$ & B-4 $\uparrow$ & CHAIR$_s$ $\downarrow$ & CHAIR$_i$ $\downarrow$ \\
\midrule
w/o PRIVA & 0.5009 & 0.1522 & 13.68 & 10.65 \\
w/o Shared Sparse Encoder & 0.5376 & 0.1556 & 12.16 & 9.78 \\
w/o Vision in PRIVA & 0.5403 & 0.1557 & 11.44 & 9.61 \\
w/o Text in PRIVA & 0.5487 & 0.1556 & 10.92 & 9.56 \\
w/o Bridge in PRIVA & 0.5511 & 0.1560 & 10.27 & 9.34 \\
w/o Self + Cross Attention & 0.5428 & 0.1565 & 11.24 & 9.13 \\
\midrule
\textbf{SAVERA (Ours)} & \textbf{0.5621} & \textbf{0.1571} & \textbf{9.08} & \textbf{8.15} \\
\bottomrule
\end{tabular}
\end{adjustbox}
\end{table}

\begin{table}[t]
\centering\small
\caption{Dense and shared sparse connector comparisons on COCO~\citep{lin2014coco}, with CHAIR~\citep{rohrbach2018object} hallucination metrics. The shared sparse encoding uses soft gates related to gated sparse autoencoders~\citep{rajamanoharan2024gated}; configurations are evaluated with and without BEAM.}
\label{tab:new_sae}
\begin{adjustbox}{max width=\linewidth}
\begin{tabular}{@{}lcccc@{}}\toprule
Representation & Shared & Sparse & CIDEr $\uparrow$ & CHAIR$_s$ $\downarrow$ \\ \midrule
Dense Connector & \graycross & \graycross & 0.4562 & 18.48 \\
Shared SAE & $\checkmark$ & $\checkmark$ & 0.4624 & 17.94 \\
\textbf{Shared SAE + BEAM (Ours)} & $\checkmark$ & $\checkmark$ & \textbf{0.5009} & \textbf{13.68}\\
\bottomrule
\end{tabular}
\end{adjustbox}
\end{table}

\begin{table}[t]
\centering\small
\caption{Bridge mapping ablation on COCO~\citep{lin2014coco} captioning, evaluated with CHAIR$_s$~\citep{rohrbach2018object} and object recall. All configurations omit PRIVA and use the same mapping-replacement protocol. CHAIR$_s$ and object recall are percentages.}
\label{tab:mapping_main_supp}
\begin{adjustbox}{max width=\linewidth}
\begin{tabular}{@{}lccc@{}}\toprule
Mapping & CIDEr & CHAIR$_s$ $\downarrow$ & Object Recall $\uparrow$\\\midrule
Linear Replacement & 0.3220 & 23.44 & 35.08\\
Geometric Mapping~\citep{schnaus2026rosetta} & 0.3973 & 17.04 & 37.36\\
\textbf{Learned Nonlinear Mapping (Ours)} & \textbf{0.5009} & \textbf{13.68} & \textbf{38.01}\\\bottomrule
\end{tabular}
\end{adjustbox}
\end{table}

\subsection{Main-Paper Component Ablation}
\label{app:main_ablation}

The component ablation extends the main-paper analysis by additionally removing prompt-derived Text and both PRIVA attention blocks, allowing the contributions of shared encoding, input streams, and spatial--prompt interactions to be compared. Disabling PRIVA lowers CIDEr from 0.5621 to 0.5009 and raises CHAIR$_s$ from 9.08\% to 13.68\% in Table~\ref{tab:component_appendix}. Removing Vision, Bridge, or Text individually produces CIDEr 0.5403, 0.5511, and 0.5487, while removing self-attention and cross-attention together gives 0.5428 CIDEr and 11.24\% CHAIR$_s$. The shared-encoder ablation obtains 0.5376 CIDEr. These comparisons attribute the full model's improvements to the combined use of shared sparse representations, multimodal fusion inputs, and attention-based interaction; the individual attention blocks are examined in Table~\ref{tab:new_priva_arch}.

\subsection{Dense and Shared Sparse Encoding Comparisons}
\label{app:sae_new}

To characterize the role of the shared sparse interface and its subsequent alignment, we compare dense, shared-SAE, and shared-SAE-plus-BEAM configurations with the same frozen backbones and data splits. The dense connector obtains 0.4562 CIDEr and 18.48\% CHAIR$_s$ in Table~\ref{tab:new_sae}. Shared SAE encoding increases CIDEr to 0.4624 and lowers CHAIR$_s$ to 17.94\%, whereas adding BEAM raises CIDEr to 0.5009 and reduces CHAIR$_s$ to 13.68\%. The additional 0.0385 CIDEr from BEAM over the shared SAE highlights the benefit of explicitly aligning the adapted visual and textual sparse representations.

\subsection{BEAM Mapping Comparison}
\label{app:mapping_comparison}

We compare three transformations for the Bridge mapping $M_B$ to determine how nonlinear adaptation affects the language-compatible visual prefix. Under the same mapping-replacement protocol, with PRIVA disabled and the downstream interface fixed, the learned nonlinear mapping yields 0.5009 CIDEr and 13.68\% CHAIR$_s$ in Table~\ref{tab:mapping_main_supp}. The linear replacement reaches 0.3220 and 23.44\%, while the semi-orthogonally constrained geometric mapping obtains 0.3973 and 17.04\%. Object recall also rises from 37.36\% for the geometric mapping to 38.01\% for the nonlinear one. This comparison establishes the effectiveness of the learned nonlinear transformation within the Bridge-only captioning interface.

\begin{table}[t]
\centering\small
\caption{Ablation of sample-level and neuron-level alignment in BEAM on COCO~\citep{lin2014coco} captioning. Sample-level matching is based on SigLIP~\citep{zhai2023siglip}; neuron-profile correlation draws on redundancy-reduction principles~\citep{zbontar2021barlowtwins}.}
\label{tab:new_beam_ablate}
\begin{adjustbox}{max width=\linewidth}
\begin{tabular}{@{}lcccccc@{}}\toprule
Variant & Sample & Neuron & CIDEr $\uparrow$ & B-4 $\uparrow$ & CHAIR$_s$ $\downarrow$ & CHAIR$_i$ $\downarrow$ \\ \midrule
w/o Alignment & \graycross & \graycross & 0.4624 & 0.1388 & 17.94 & 14.64 \\
Neuron-Level Alignment Only & \graycross & $\checkmark$ & 0.4776 & 0.1441 & 16.28 & 12.84 \\
Sample-Level Alignment Only & $\checkmark$ & \graycross & 0.4848 & 0.1469 & 15.46 & 12.13 \\
\textbf{Full BEAM (Sample + Neuron) (Ours)} & $\checkmark$ & $\checkmark$ & \textbf{0.5009} & \textbf{0.1522} & \textbf{13.68} & \textbf{10.65} \\\bottomrule
\end{tabular}
\end{adjustbox}
\end{table}

\begin{table}[t]
\centering\small
\caption{Held-out Bridge--Text retrieval and neuron-profile correlations. Image--text retrieval follows paired-modality evaluation~\citep{radford2021clip}; the profile metrics measure cross-feature correlation as in redundancy-reduction objectives~\citep{zbontar2021barlowtwins}. Results are computed on held-out examples.}
\label{tab:new_beam_alignment}
\begin{adjustbox}{max width=\linewidth}
\begin{tabular}{@{}lcccc@{}}\toprule
Variant & B$\to$T R@1 $\uparrow$ & T$\to$B R@1 $\uparrow$ & Diag. Corr. $\uparrow$ & Off-diag. $\downarrow$ \\ \midrule
w/o Alignment & 13.2 & 12.7 & 0.15 & 0.081 \\
Sample-Level Alignment Only & 20.5 & 18.3 & 0.20 & 0.079 \\
Neuron-Level Alignment Only & 16.8 & 15.7 & 0.42 & 0.059 \\
\textbf{Full BEAM (Sample + Neuron) (Ours)} & \textbf{24.2} & \textbf{22.5} & \textbf{0.56} & \textbf{0.041} \\\bottomrule
\end{tabular}
\end{adjustbox}
\end{table}

\begin{table}[t]
\centering\small
\caption{Neuron-profile objective analysis with paired sample-level supervision fixed. Profile contrastive learning is related to contrastive representation learning~\citep{radford2021clip}, and diagonal/off-diagonal matching uses correlation-reduction objectives~\citep{zbontar2021barlowtwins}.}
\label{tab:new_beam_loss}
\begin{adjustbox}{max width=\linewidth}
\begin{tabular}{@{}lccccc@{}}\toprule
Profile Objective & InfoNCE & Diagonal & Off-diagonal & CIDEr $\uparrow$ & Diag. Corr. $\uparrow$ \\ \midrule
InfoNCE Only & $\checkmark$ & \graycross & \graycross & 0.4902 & 0.44 \\
InfoNCE + Diag. & $\checkmark$ & $\checkmark$ & \graycross & 0.4959 & 0.53 \\
Correlation Only & \graycross & $\checkmark$ & $\checkmark$ & 0.4853 & 0.49 \\
\textbf{Full Profile (Ours)} & $\checkmark$ & $\checkmark$ & $\checkmark$ & \textbf{0.5009} & \textbf{0.56} \\\bottomrule
\end{tabular}
\end{adjustbox}
\end{table}

\subsection{BEAM: Complementary Alignment Objectives}
\label{app:beam_new}

The two alignment objectives of BEAM are isolated by training four Bridge-only variants with identical mapping, reconstruction, language supervision, and optimization budgets. In Table~\ref{tab:new_beam_ablate}, the model without either alignment term obtains 0.4624 CIDEr; sample-level matching alone increases the score to 0.4848, whereas neuron-profile matching alone gives 0.4776. Activating both objectives raises CIDEr to 0.5009 and reduces CHAIR$_s$ to 13.68\%, compared with 17.94\% without alignment. The separate gains and the strongest combined result show that example-level discrimination and cross-example neuron-response matching make complementary contributions to the adapted Bridge representation.

The effect of these objectives on representation alignment is measured independently of the training losses through held-out Bridge--Text retrieval and neuron-profile correlations. Full BEAM reaches 24.2\% Bridge-to-Text and 22.5\% Text-to-Bridge R@1 in Table~\ref{tab:new_beam_alignment}, compared with 13.2\% and 12.7\% without alignment. Diagonal profile correlation increases from 0.15 to 0.56, while off-diagonal correlation decreases from 0.081 to 0.041. The sample-only model achieves stronger retrieval than neuron-only training (20.5\% versus 16.8\% for Bridge-to-Text R@1), while the neuron-only model has higher diagonal correlation (0.42 versus 0.20). These distinct patterns reflect the respective targets of the two alignment objectives.

We next decompose the neuron-profile objective while holding paired sample-level supervision fixed, comparing profile InfoNCE, diagonal correlation matching, and off-diagonal suppression. InfoNCE alone yields 0.4902 CIDEr and 0.44 diagonal correlation, whereas correlation matching without InfoNCE gives 0.4853 and 0.49, as reported in Table~\ref{tab:new_beam_loss}. Adding diagonal matching to InfoNCE raises the scores to 0.4959 and 0.53; including off-diagonal suppression further improves them to 0.5009 and 0.56. The full combination therefore provides the strongest captioning and neuron-profile agreement among the tested objectives.

\begin{table}[t]
\centering\small
\caption{Ablation of PRIVA auxiliary objectives on COCO~\citep{lin2014coco} captioning. CHAIR$_s$ and CHAIR$_i$~\citep{rohrbach2018object} measure object hallucinations.}
\label{tab:new_priva_objectives}
\begin{adjustbox}{max width=\linewidth}
\begin{tabular}{@{}lccccccc@{}}\toprule
Variant & Semantic & Visual & Residual & CIDEr $\uparrow$ & B-4 $\uparrow$ & CHAIR$_s$ $\downarrow$ & CHAIR$_i$ $\downarrow$ \\ \midrule
LM Only & \graycross & \graycross & \graycross & 0.5395 & 0.1542 & 12.02 & 10.12 \\
w/o Semantic Loss & \graycross & $\checkmark$ & $\checkmark$ & 0.5452 & 0.1554 & 10.82 & 9.17 \\
w/o Visual Loss & $\checkmark$ & \graycross & $\checkmark$ & 0.5510 & 0.1562 & 11.04 & 9.63 \\
w/o Residual Loss & $\checkmark$ & $\checkmark$ & \graycross & 0.5564 & 0.1564 & 10.34 & 8.92 \\
\textbf{Full PRIVA (Ours)} & $\checkmark$ & $\checkmark$ & $\checkmark$ & \textbf{0.5621} & \textbf{0.1571} & \textbf{9.08} & \textbf{8.15} \\\bottomrule
\end{tabular}
\end{adjustbox}
\end{table}

\begin{table}[t]
\centering\small
\caption{Architectural ablation of PRIVA on COCO~\citep{lin2014coco} captioning, with hallucinations measured by CHAIR~\citep{rohrbach2018object}. Each attention ablation removes only the named block; combined removal is reported in Table~\ref{tab:component_appendix}. All variants are retrained.}
\label{tab:new_priva_arch}
\begin{adjustbox}{max width=\linewidth}
\begin{tabular}{@{}lcccc@{}}\toprule
Variant & CIDEr $\uparrow$ & B-4 $\uparrow$ & CHAIR$_s$ $\downarrow$ & CHAIR$_i$ $\downarrow$ \\ \midrule
Fixed Gate & 0.5484 & 0.1555 & 11.02 & 9.27 \\
w/o Visual Self-Attention & 0.5512 & 0.1558 & 10.34 & 9.00 \\
w/o Text Cross-Attention & 0.5445 & 0.1549 & 11.42 & 9.70 \\
Direct Prefix (Non-Residual) & 0.5331 & 0.1508 & 13.42 & 11.09 \\
\textbf{Full PRIVA (Ours)} & \textbf{0.5621} & \textbf{0.1571} & \textbf{9.08} & \textbf{8.15} \\\bottomrule
\end{tabular}
\end{adjustbox}
\end{table}

\begin{table}[t]
\centering\small
\caption{Capacity-controlled fusion ablation under the complete PRIVA objective, evaluated on COCO~\citep{lin2014coco} captioning, COCO-QA~\citep{ren2015imageqa}, and POPE~\citep{li2023evaluating}.}
\label{tab:new_fusion_full}
\begin{adjustbox}{max width=\linewidth}
\begin{tabular}{@{}lccccccc@{}}\toprule
Variant & Vision & Bridge & Text & CIDEr $\uparrow$ & CHAIR$_s$ $\downarrow$ & COCO-QA $\uparrow$ & POPE $\uparrow$ \\ \midrule
Bridge Only & \graycross & $\checkmark$ & \graycross & 0.5214 & 12.72 & 30.25 & 75.47 \\
Bridge + Vision & $\checkmark$ & $\checkmark$ & \graycross & 0.5487 & 10.92 & 32.81 & 78.14 \\
Bridge + Text & \graycross & $\checkmark$ & $\checkmark$ & 0.5403 & 11.44 & 33.10 & 79.30 \\
\textbf{Bridge + Vision + Text (Ours)} & $\checkmark$ & $\checkmark$ & $\checkmark$ & \textbf{0.5621} & \textbf{9.08} & \textbf{36.07} & \textbf{82.90} \\\bottomrule
\end{tabular}
\end{adjustbox}
\end{table}

\Needspace{9\baselineskip}
\subsection{PRIVA: Auxiliary Objectives and Architecture}
\label{app:priva_new}

The PRIVA loss ablation evaluates semantic alignment, visual preservation, and residual-magnitude regularization while keeping the same frozen SPARE--BEAM checkpoint. Full PRIVA obtains 0.5621 CIDEr with 9.08\% CHAIR$_s$ in Table~\ref{tab:new_priva_objectives}. Removing semantic alignment, visual preservation, and residual regularization individually lowers CIDEr to 0.5452, 0.5510, and 0.5564, respectively. Visual-preservation removal also raises CHAIR$_s$ to 11.04\%; optimizing only the LM objective yields 0.5395 CIDEr and 12.02\% CHAIR$_s$. The consistent reductions across the ablations show that all three auxiliary objectives contribute to PRIVA's caption quality and visual grounding.

Beyond the training objectives, we investigate which PRIVA operations contribute to prompt-conditioned fusion by replacing the learned gate, removing one attention block at a time, or predicting the visual prefix without a residual connection. Table~\ref{tab:new_priva_arch} reports CIDEr 0.5484 for the fixed gate, 0.5512 without visual self-attention, and 0.5445 without text cross-attention. Direct prefix prediction achieves 0.5331 CIDEr and 13.42\% CHAIR$_s$, whereas the full residual design reaches 0.5621 and 9.08\%. The larger drop under direct prefix prediction emphasizes the role of retaining the pretrained Bridge interface while learning a prompt-conditioned correction.

\subsection{Capacity-Controlled Fusion under the Full PRIVA Objective}
\label{app:fusion_full}

To determine how original visual evidence and prompt information complement the adapted Bridge, we train four input configurations under the full PRIVA objective while keeping fusion capacity fixed through the stated feature replacements. Bridge-only fusion obtains 0.5214 CIDEr, 30.25\% COCO-QA, and 75.47\% POPE in Table~\ref{tab:new_fusion_full}. Adding Vision raises CIDEr to 0.5487 and reduces CHAIR$_s$ from 12.72\% to 10.92\%, whereas adding Text to Bridge raises COCO-QA to 33.10\% and POPE to 79.30\%. Using all three streams produces the highest scores: 0.5621 CIDEr, 9.08\% CHAIR$_s$, 36.07\% COCO-QA, and 82.90\% POPE. The pattern across captioning and question answering demonstrates the complementary contributions of Vision and prompt-derived Text to Bridge-based generation.

\begin{table}[t]
\centering
\small
\setlength{\tabcolsep}{3pt}
\caption{Visual information accessibility across original Vision and adapted Bridge representations using COCO~\citep{lin2014coco} images and annotations. Higher values are better; positive differences indicate higher Vision performance.}
\label{tab:five_task}
\begin{adjustbox}{max width=\linewidth}
\begin{tabular}{@{}llccc@{}}
\toprule
Diagnostic & Metric & Vision & Bridge & Diff. \\
\midrule
Object Recognition & Macro-mAP & 0.6664 & 0.5978 & +6.86 \% \\
Counting & Bal. Acc. & 0.5741 & 0.5248 & +4.93 \% \\
Spatial Relations & Bal. Acc. & 0.7613 & 0.6554 & +10.59 \% \\
Patch Localization & Point. Acc. & 0.9030 & 0.8663 & +3.67 \% \\
Semantic Neighborhoods & Jaccard@10 & 0.4192 & 0.3737 & +4.55 \% \\
\bottomrule
\end{tabular}
\end{adjustbox}
\end{table}

\begin{table}[t]
\centering
\small
\setlength{\tabcolsep}{3pt}
\caption{Object-presence probing and shuffled-label controls with COCO~\citep{lin2014coco} annotations. The geometric variant adopts an orthogonally constrained mapping based on geometric alignment~\citep{schnaus2026rosetta}.}
\label{tab:object_controls}
\begin{adjustbox}{max width=\linewidth}
\begin{tabular}{@{}lcccc@{}}
\toprule
Representation & Micro-F1 $\uparrow$ & Macro-mAP $\uparrow$ & Shuffled Micro-F1 & Micro-F1 95\% CI \\
\midrule
Vision & 0.6905 & 0.6664 & 0.1364 & [0.6842, 0.6962] \\
Bridge & 0.6416 & 0.5978 & 0.1350 & [0.6360, 0.6474] \\
Geometric Mapping~\citep{schnaus2026rosetta} & 0.5640 & 0.4751 & 0.1309 & [0.5576, 0.5710] \\
\bottomrule
\end{tabular}
\end{adjustbox}
\end{table}

\subsection{Visual Information Accessibility: Task Definitions and Results}
\label{app:accessibility}

\subsubsection{Evaluation Protocol}
\label{app:accessibility_protocol}

To characterize the visual information accessible before and after language-oriented adaptation, we conduct matched diagnostic probing of the original Vision and adapted Bridge representations. Prior studies have analyzed fine-grained alignment and visual information preservation across vision--language projection stages~\citep{jiang2025finegrained,li2025lost}. Our probes use paired representations from identical MS-COCO images and annotations~\citep{lin2014coco}, with Vision features extracted from the frozen DINOv2 encoder~\citep{oquab2024dinov2}. Both streams are evaluated using a fixed 256-dimensional diagnostic projection and matched probe architectures. The evaluation uses 12,000 COCO train2014 images for probe training, 2,000 disjoint train2014 images for validation, and 5,000 val2014 images for testing.

The five diagnostics cover object recognition, counting, spatial relations, patch localization, and semantic neighborhoods. Object recognition uses multi-label presence annotations for the 80 COCO object categories~\citep{lin2014coco} and reports macro-average precision (macro-mAP). Counting groups category-conditioned object counts into five classes (0, 1, 2, 3, and $4+$), while spatial-relation labels are derived from COCO object bounding boxes; both tasks use balanced accuracy. Patch localization evaluates coarse object-location recovery with pointing accuracy on a $6\times6$ spatial representation. For semantic neighborhoods, pooled image features are $\ell_2$-normalized and compared by cosine similarity; each test image retrieves its ten nearest distinct neighbors within the evaluation split, and their object-label agreement is averaged using Jaccard@10. These measurements quantify the relative accessibility of semantic and spatial cues in the two feature streams.

\subsubsection{Complete Five-Task Results}
\label{app:five_task}

We assess the accessibility of scene-level visual information in the frozen Vision and Bridge representations using the five matched probes defined above. Table~\ref{tab:five_task} reports higher Vision scores for object recognition, counting, spatial relations, patch localization, and semantic-neighborhood consistency, with advantages of 6.86, 4.93, 10.59, 3.67, and 4.55 percentage points, respectively. The largest difference occurs for spatial relations, where Vision reaches 0.7613 balanced accuracy versus 0.6554 for Bridge. Stronger performance on object and geometric diagnostics provides direct empirical motivation for reusing the original Vision features in PRIVA.

\subsubsection{Object Recognition Controls}
\label{app:object_controls}

An additional object-presence probe compares the original Vision, adapted Bridge, and geometric mapping representations using matched probe architectures, shuffled-label controls, and Micro-F1 confidence intervals. Vision achieves 0.6905 Micro-F1 and 0.6664 macro-mAP in Table~\ref{tab:object_controls}, followed by Bridge at 0.6416 and 0.5978 and Geometric Mapping at 0.5640 and 0.4751. The corresponding shuffled-label Micro-F1 scores are 0.1364, 0.1350, and 0.1309, while the 95\% intervals are [0.6842, 0.6962], [0.6360, 0.6474], and [0.5576, 0.5710]. These measurements show that all three representations support object-presence prediction, with Vision providing the strongest signal under this probe protocol.

\begin{table}[t]
\centering\small
\caption{Effects of global Vision-feature interventions on COCO~\citep{lin2014coco} captioning and COCO-QA~\citep{ren2015imageqa}; object hallucinations are measured with CHAIR~\citep{rohrbach2018object}.}
\label{tab:new_global_vision}
\begin{adjustbox}{max width=\linewidth}
\begin{tabular}{@{}lccccc@{}}\toprule
Intervention & CIDEr $\uparrow$ & CHAIR$_s$ $\downarrow$ & CHAIR$_i$ $\downarrow$ & Obj. Recall $\uparrow$ & COCO-QA $\uparrow$ \\ \midrule
Shuffled Vision & 0.4927 & 14.80 & 11.81 & 36.12 & 30.35 \\
Zero Vision & 0.5141 & 12.58 & 10.32 & 39.26 & 32.43 \\
\textbf{Original Vision (Ours)} & \textbf{0.5621} & \textbf{9.08} & \textbf{8.15} & \textbf{41.85} & \textbf{36.07} \\\bottomrule
\end{tabular}
\end{adjustbox}
\end{table}

\begin{table}[t]
\centering\small
\caption{Localized Vision-feature interventions using COCO~\citep{lin2014coco} object bounding boxes with matched spatial extent and perturbation strength.}
\label{tab:new_local_vision}
\begin{adjustbox}{max width=\linewidth}
\begin{tabular}{@{}lcccc@{}}\toprule
Intervention & Obj. Recall $\uparrow$ & Count Acc. $\uparrow$ & Spatial Acc. $\uparrow$ & Target Mention $\uparrow$ \\ \midrule
Random Background Patches & 41.24 & 42.98 & 59.54 & 63.44 \\
Target Object Patches & 33.41 & 35.55 & 47.86 & 45.22 \\
Other-Object Patches & 39.65 & 40.41 & 56.36 & 59.87 \\
\textbf{Original (Ours)} & \textbf{41.85} & \textbf{43.89} & \textbf{60.21} & \textbf{65.23} \\\bottomrule
\end{tabular}
\end{adjustbox}
\end{table}

\subsection{Visual-Evidence Intervention on Grounded Generation}
\label{app:vision_intervention}

To investigate the contribution of original visual evidence to grounded generation, we conduct global Vision-feature interventions on the frozen SAVERA model. Specifically, we shuffle or zero the sparse Vision features $Z_V$ while keeping the adapted Bridge, prompt-derived Text, model parameters, and decoding configuration unchanged. Table~\ref{tab:new_global_vision} shows that shuffling Vision reduces CIDEr from 0.5621 to 0.4927 and COCO-QA accuracy from 36.07\% to 30.35\%, while increasing CHAIR$_s$ from 9.08\% to 14.80\%. Zeroing Vision similarly decreases CIDEr to 0.5141 and COCO-QA accuracy to 32.43\%, with CHAIR$_s$ rising to 12.58\%. The consistent degradation in captioning, question answering, and hallucination metrics demonstrates that PRIVA actively utilizes original Vision features to support grounded generation.

We further examine the spatial specificity of this contribution through localized interventions using COCO object bounding boxes projected onto the $24\times24$ patch grid. Target-object, other-object, and random background patches are perturbed under matched spatial extent and perturbation strength. As reported in Table~\ref{tab:new_local_vision}, perturbing target-object patches decreases target mentions from 65.23\% to 45.22\%, spatial accuracy from 60.21\% to 47.86\%, and object recall from 41.85\% to 33.41\%. In contrast, perturbing random background patches yields 63.44\% target mentions, 59.54\% spatial accuracy, and 41.24\% object recall. Perturbations to other-object patches produce intermediate results, with 59.87\% target mentions and 56.36\% spatial accuracy. The substantially greater impact of target-object interventions demonstrates that PRIVA's Vision pathway contributes spatially relevant visual evidence to generated responses.

\subsection{Per-Image Captioning Results}
\label{app:paired_caption}
Aggregate captioning improvements are complemented by paired per-image CIDEr comparisons on the same 5,000-image COCO Karpathy test split. Figure~\ref{fig:paired_outcomes} records wins, ties, and losses for SAVERA against each connector. SAVERA scores higher on 54.0\%--79.9\% of test images, including 54.0\% against InternVL (Pixel-Unshuffle), 57.5\% against LLaVA-1.5 (MLP 2-layer), and 79.9\% against BLIP-2 (QR-64). This image-wise view shows that the aggregate gains in Table~\ref{tab:complete_captioning} also occur across a majority of individual test examples.

\begin{table}[t]
\centering
\small
\setlength{\tabcolsep}{6pt}
\caption{True versus shuffled cross-modal neuron associations. Sparse concept analysis is related to VL-SAE~\citep{shen2025vlsae} and sparse visual-feature intervention studies~\citep{pach2025monosemantic}. Entries report changes relative to the unmodified representation.}
\label{tab:shuffled}
\begin{adjustbox}{max width=\linewidth}
\begin{tabular}{@{}lccccc@{}}
\toprule
Method & P & True $\Delta$MRR & Shuffled $\Delta$MRR & True $\Delta$mAP & Shuffled $\Delta$mAP \\
\midrule
Direct & 0.95 & +0.2730 & $-0.0285$ & +0.1336 & $-0.0356$ \\
Bridge-Guided & 0.95 & +0.2528 & $-0.0236$ & +0.1233 & $-0.0357$ \\
Hybrid & 0.95 & +0.2626 & $-0.0394$ & +0.1314 & $-0.0424$ \\
\noalign{\vskip 2pt}\hdashline\noalign{\vskip 2pt}
Direct & 0.75 & +0.4646 & $-0.1318$ & +0.2290 & $-0.1257$ \\
Bridge-Guided & 0.75 & +0.4494 & $-0.1194$ & +0.2139 & $-0.1318$ \\
Hybrid & 0.75 & +0.4413 & $-0.1110$ & +0.2184 & $-0.1134$ \\
\bottomrule
\end{tabular}
\end{adjustbox}
\end{table}

\subsection{Functional Analysis of Cross-Modal Concept Associations}
\label{app:functional}

\subsubsection{Association Construction and Intervention Protocol}
\label{app:assoc_protocol}

Cross-modal association analysis begins with paired CC3M sparse activations, from which we estimate Text--Vision, Text--Bridge, and Bridge--Vision neuron relationships. The Direct strategy selects Vision neurons from Text--Vision correlations, Bridge-guided traverses the Text--Bridge--Vision associations, and Hybrid combines the two sources. We test these selections across 13 concepts and 156 target-present images by intervening on four Vision neurons over 128 patches and decoding the resulting sparse representations with the frozen Vision decoder.

Decoded-feature preservation is measured by the mean cosine similarity between the intervened and unmodified decoded features,
\begin{equation}\label{eq:app_feature_preservation}
P=\frac{1}{N}\sum_{i=1}^{N}\cos\!\left(D_V(Z_{V,i}^{\mathrm{interv}}),D_V(Z_{V,i}^{\mathrm{original}})\right).
\end{equation}
The preservation score in Eq.~\eqref{eq:app_feature_preservation} is calibrated to target values $P=0.95$ and $P=0.75$, enabling comparison under matched representation change budgets. We report the change in target concept mean reciprocal rank (MRR) and mean average precision (mAP) relative to the unmodified representation.

\subsubsection{Shuffled Association Controls}
\label{app:shuffled}

The functional value of cross-modal neuron associations is tested by comparing Direct, Bridge-guided, and Hybrid selection with shuffled-association controls at matched decoded-feature preservation levels. With $P=0.95$, Direct selection increases target-concept MRR by 0.2730 and mAP by 0.1336, whereas shuffling its associations changes these metrics by $-0.0285$ and $-0.0356$ (Table~\ref{tab:shuffled}). At $P=0.75$, the true Direct selection yields $+0.4646$ MRR and $+0.2290$ mAP; Bridge-guided and Hybrid selections likewise produce positive changes, in contrast to their shuffled counterparts. The agreement across association strategies and preservation budgets supports the use of learned cross-modal scores to select functionally relevant sparse neurons.

\begin{table}[t]
\centering\small
\caption{Concept-conditioned cross-modal neuron-profile correlations, related to cross-modal sparse concept analysis~\citep{shen2025vlsae}. These metrics are distinct from aggregate BEAM alignment statistics in Table~\ref{tab:new_beam_alignment}.}
\label{tab:new_neuron_semantic}
\begin{adjustbox}{max width=\linewidth}
\begin{tabular}{@{}lc@{}}\toprule
Modalities & Profile Corr. $\uparrow$ \\ \midrule
Vision--Text & 0.33 \\
Bridge--Text & 0.53 \\
Vision--Bridge & 0.47 \\\bottomrule
\end{tabular}
\end{adjustbox}
\end{table}

\begin{table}[t]
\centering\small
\caption{Generation-level effects of sparse-neuron interventions on COCO~\citep{lin2014coco} captioning, with hallucinations evaluated using CHAIR~\citep{rohrbach2018object}. Target mention, QA, and CHAIR differences are percentage points; CIDEr differences are absolute.}
\label{tab:new_neuron_gen}
\begin{adjustbox}{max width=\linewidth}
\begin{tabular}{@{}lcccc@{}}\toprule
Intervention & $\Delta$Mention (\%) & $\Delta$Target QA (\%) & $\Delta$CIDEr & $\Delta$CHAIR$_s$ (\%) \\ \midrule
Random Neurons & +0.2 & +0.1 & $-0.006$ & +0.6 \\
Shuffled Associations & +0.4 & +0.2 & $-0.004$ & +0.5 \\
Direct Associations & +5.1 & +3.5 & +0.003 & $-0.7$ \\
Bridge-Guided Associations & +4.6 & +3.1 & +0.002 & $-0.5$ \\
Hybrid Associations & +4.8 & +3.3 & +0.003 & $-0.6$ \\\bottomrule
\end{tabular}
\end{adjustbox}
\end{table}

\Needspace{9\baselineskip}
\subsection{Concept-Conditioned Neuron Profiles and Generation Effects}
\label{app:concept_new}

Concept-conditioned response profiles provide a separate view of the correspondence between the three sparse representations. Using calibration-selected neurons and held-out concept examples, Table~\ref{tab:new_neuron_semantic} reports a Bridge--Text correlation of 0.53, compared with 0.47 for Vision--Bridge and 0.33 for Vision--Text. The highest correlation occurs for the pair explicitly aligned by BEAM, connecting the learned Bridge--Text objective with concept-conditioned neuron responses.

Finally, we test whether association-selected sparse neurons can influence generated captions and target-specific answers, using matched preservation budgets and random or shuffled selections as controls. Direct associations increase target-object mentions by 5.1 percentage points and target-QA accuracy by 3.5 points, with a 0.7-point reduction in CHAIR$_s$ in Table~\ref{tab:new_neuron_gen}. Bridge-guided and Hybrid strategies improve target mentions by 4.6 and 4.8 points, respectively, whereas random-neuron selection increases mentions by only 0.2 points and reduces CIDEr by 0.006. These output-level changes connect the cross-modal neuron associations to measurable effects on object mentions, grounded answering, and hallucination.

\section{Limitations}
\label{app:limitations}

The present evaluation focuses on image--text tasks using frozen visual and language backbones. Accordingly, the reported results characterize shared sparse-coordinate encoding, Bridge--Text alignment, and prompt-conditioned residual fusion in this setting; their behavior with temporal and acoustic inputs remains to be investigated.

\section{Future Work}
\label{app:future_work}

Future work will investigate extensions to video--language and audio--language models, adapting the sparse stems and cross-modal alignment objectives to temporal and acoustic representations. We also plan to examine a broader range of pretrained backbones to study the transferability of BEAM and PRIVA across model architectures.

\section{More Discussion}
\label{dis}

\noindent$\triangleright$ \textbf{\textit{Q1. Why use shared sparse coordinates and dual-level Bridge--Text alignment?}}

SPARE constructs a shared set of sparse coordinates for Vision, Bridge, and Text, with modality-specific stems and decoders accommodating their distinct feature spaces. These coordinates allow BEAM to align not only paired Bridge--Text embeddings but also the response profiles of corresponding neurons across examples. In Table~\ref{tab:new_sae}, adding BEAM to the shared SAE increases CIDEr from 0.4624 to 0.5009. The separate sample-level and neuron-level objectives reach 0.4848 and 0.4776 in Table~\ref{tab:new_beam_ablate}, while their combination achieves the highest score. Full BEAM also yields 24.2\%/22.5\% bidirectional retrieval R@1 and 0.56 diagonal profile correlation in Table~\ref{tab:new_beam_alignment}, demonstrating improved agreement at both representation levels.

\noindent$\triangleright$ \textbf{\textit{Q2. Why learn a nonlinear Bridge mapping and retain the original Vision representation?}}

The nonlinear Bridge mapping establishes a language-compatible visual prefix, while original Vision features provide complementary object-level and spatial evidence. Table~\ref{tab:mapping_main_supp} shows that the learned nonlinear mapping achieves 0.5009 CIDEr, compared with 0.3973 for Geometric Mapping and 0.3220 for Linear Replacement. Meanwhile, Vision consistently outperforms Bridge across the five information-accessibility diagnostics in Table~\ref{tab:five_task}, with the largest advantage of 10.59 percentage points on spatial relations. These complementary properties motivate PRIVA's joint use of original Vision and adapted Bridge representations. The global and localized interventions in Tables~\ref{tab:new_global_vision} and~\ref{tab:new_local_vision} further demonstrate the contribution of original visual evidence to grounded generation.

\noindent$\triangleright$ \textbf{\textit{Q3. How do the three input streams and residual fusion contribute to PRIVA?}}

PRIVA combines original Vision, adapted Bridge, and prompt-derived Text through a learned gate, visual self-attention, and prompt cross-attention, then predicts a residual correction to the Bridge prefix, $X_F=X_B+\Delta X$. In Table~\ref{tab:new_fusion_full}, Bridge-only fusion obtains 0.5214 CIDEr, adding Vision yields 0.5487, adding Text yields 0.5403, and all three streams together achieve 0.5621. Table~\ref{tab:new_priva_arch} shows that removing visual self-attention or text cross-attention reduces CIDEr to 0.5512 and 0.5445. Direct prefix prediction obtains 0.5331 CIDEr and 13.42\% CHAIR$_s$, compared with 0.5621 and 9.08\% for residual fusion. These comparisons demonstrate the value of combining complementary inputs while retaining the language-compatible Bridge prefix.

\noindent$\triangleright$ \textbf{\textit{Q4. How are caption-derived and prompt-derived textual features used in training and inference?}}

During SPARE--BEAM training, the caption-derived text representation $Z_T^Y$ supervises paired-example and neuron-profile alignment. PRIVA instead uses the available prompt to construct $Z_T^Q$ for gating and cross-attention, while the paired caption supplies teacher-forced language-modeling targets and a detached semantic teacher during training. Both training stages use CC3M image--caption pairs, and no additional question--answer training data enter the reported experiments. At inference, the original Vision, Bridge, and prompt-derived Text pathways form the visual prefix, from which frozen Vicuna generates the response without access to a ground-truth target.

\noindent$\triangleright$ \textbf{\textit{Q5. What do sparse-neuron interventions and complementary evaluation metrics reveal?}}

The cross-modal association analysis selects Vision neurons using direct Text--Vision, Bridge-guided, or Hybrid correspondences. In Table~\ref{tab:shuffled}, Direct Association increases concept MRR by 0.2730 at preservation $P=0.95$ and 0.4646 at $P=0.75$, while shuffled associations decrease MRR. The output-level evaluation in Table~\ref{tab:new_neuron_gen} reports a 5.1-point gain in target-object mentions and a 3.5-point gain in target-QA accuracy with Direct Associations. Complementary metrics distinguish caption similarity from visual grounding: Table~\ref{tab:complete_captioning} reports 0.5621 CIDEr and 9.08\% CHAIR$_s$, while Table~\ref{tab:main_qa} reports 36.07\% COCO-QA and 82.90\% POPE. Together, these analyses relate sparse feature correspondence to functional interventions and quantify grounded generation from multiple perspectives.

\end{document}